\pdfoutput=1

\documentclass[11pt]{article}

\usepackage{acl}

\usepackage{times}
\usepackage{latexsym}

\usepackage[T1]{fontenc}

\usepackage[utf8]{inputenc}

\usepackage{microtype}

\usepackage{inconsolata}

\usepackage{CJKutf8}
\usepackage{multirow}
\usepackage{multirow}
\usepackage{booktabs}
\usepackage{graphicx}
\usepackage{makecell}
\usepackage{bbding}
\usepackage{color}
\usepackage{tabularx}

\title{Understanding the Therapeutic Relationship between Counselors and Clients in Online Text-based Counseling using LLMs}

\author{\textbf{Anqi Li}$^{1,2,3}$, \textbf{Yu Lu}$^{2}$, \textbf{Nirui Song}$^{2}$,  \textbf{Shuai Zhang}$^{1,2,3}$, \textbf{Lizhi Ma}$^{2,4,5}$\thanks{\ \ Work done as a post-doctoral researcher at Westlake University.} 
 \footnotemark[2], \textbf{Zhenzhong Lan}$^{2,3}$\thanks{\ \ Corresponding Author.} \\
$^{1}$ Zhejiang University \\ 
$^{2}$ School of Engineering, Westlake University \\
$^{3}$ Westlake University Research Center for Industries of the Future \\
$^{4}$ Zhejiang Philosophy and Social Science Laboratory \\ for Research in Early Development and Childcare \\
$^{5}$ Department of Psychology, Jing Hengyi School of Education, Hangzhou Normal University \\
\texttt{\{lianqi, malizhi, lanzhenzhong\}@westlake.edu.cn} \\
}

\begin{document}
\begin{CJK*}{UTF8}{gbsn}

\maketitle
\begin{abstract}

Robust therapeutic relationships between counselors and clients are fundamental to counseling effectiveness. The assessment of therapeutic alliance is well-established in traditional face-to-face therapy but may not directly translate to text-based settings. With millions of individuals seeking support through online text-based counseling, understanding the relationship in such contexts is crucial.

In this paper, we present an automatic approach using large language models (LLMs) to understand the development of therapeutic alliance in text-based counseling. We adapt a theoretically grounded framework specifically to the context of online text-based counseling and develop comprehensive guidelines for characterizing the alliance. We collect a comprehensive counseling dataset and conduct multiple expert evaluations on a subset based on this framework. Our LLM-based approach, combined with guidelines and simultaneous extraction of supportive evidence underlying its predictions, demonstrates effectiveness in identifying the therapeutic alliance. Through further LLM-based evaluations on additional conversations, our findings underscore the challenges counselors face in cultivating strong online relationships with clients. Furthermore, we demonstrate the potential of LLM-based feedback mechanisms to enhance counselors' ability to build relationships, supported by a small-scale proof-of-concept.

\end{abstract}

\section{Introduction}
Globally, approximately one in five individuals experience mental health problems each year~\citep{eysenbach2004health, steel2014global, holmes2018lancet}. Owing to the high costs and geographical limitations associated with traditional face-to-face therapy, coupled with concerns about stigma~\citep{white2001stigma}, many individuals are turning to seek support through online text-based psychological counseling~\citep{rochlen2004online, hanley2009internet_counselling}. However, in real-world scenarios of such counseling approaches, it remains largely unknown whether counselors and clients have established strong therapeutic alliances solely through textual communications.

In psychological counseling, a positive relationship between counselors and clients is fundamental for achieving effective therapeutic outcomes~\citep{tichenor1989comparison_wa, horvath1991relation, knaevelsrud2006wa_outcome_online}. The robust therapeutic alliance signifies the cooperative relationship between counselors and clients, characterized by their shared therapeutic goals and their ability to engage together, within the context of an affective bond or positive attachment~\citep{constantino2002working}. The absence of regular and timely assessment and feedback may impede counselors in gradually nurturing of a solid rapport with clients, potentially leading to client dropout. Traditionally, counselors rely on professional supervisors for feedback, a process that is often labor-intensive and time-consuming, resulting in delayed responses. Natural Language Processing (NLP) systems capable of understanding the alliance could empower counselors with prompt and insightful feedback to enhance their practice. However, the current understanding of the alliance is primarily limited to traditional face-to-face, speech-based counseling~\citep{martinez2019personae_wa, goldberg2020ml_wa, lin2022transformer_wai}, due to the scarcity of resources and methods tailored to text-based interactions. Also, while previous NLP research has focused on the behaviors of individual participants, either counselors' strategies~\citep{cao2019observing, gibson2016deep, chiu2024computational} or clients' reactions~\citep{tanana2015recursive, li2022towards, li-etal-2023-understanding}, the mutual cognitive and affective agreement they reach through multi-turn interactions is the key factor influencing conversation outcomes~\citep{rector1999cognitive, watson2005relation}. 

\begin{figure*}[h]
\scalebox{0.62}{
    \centering
    \includegraphics{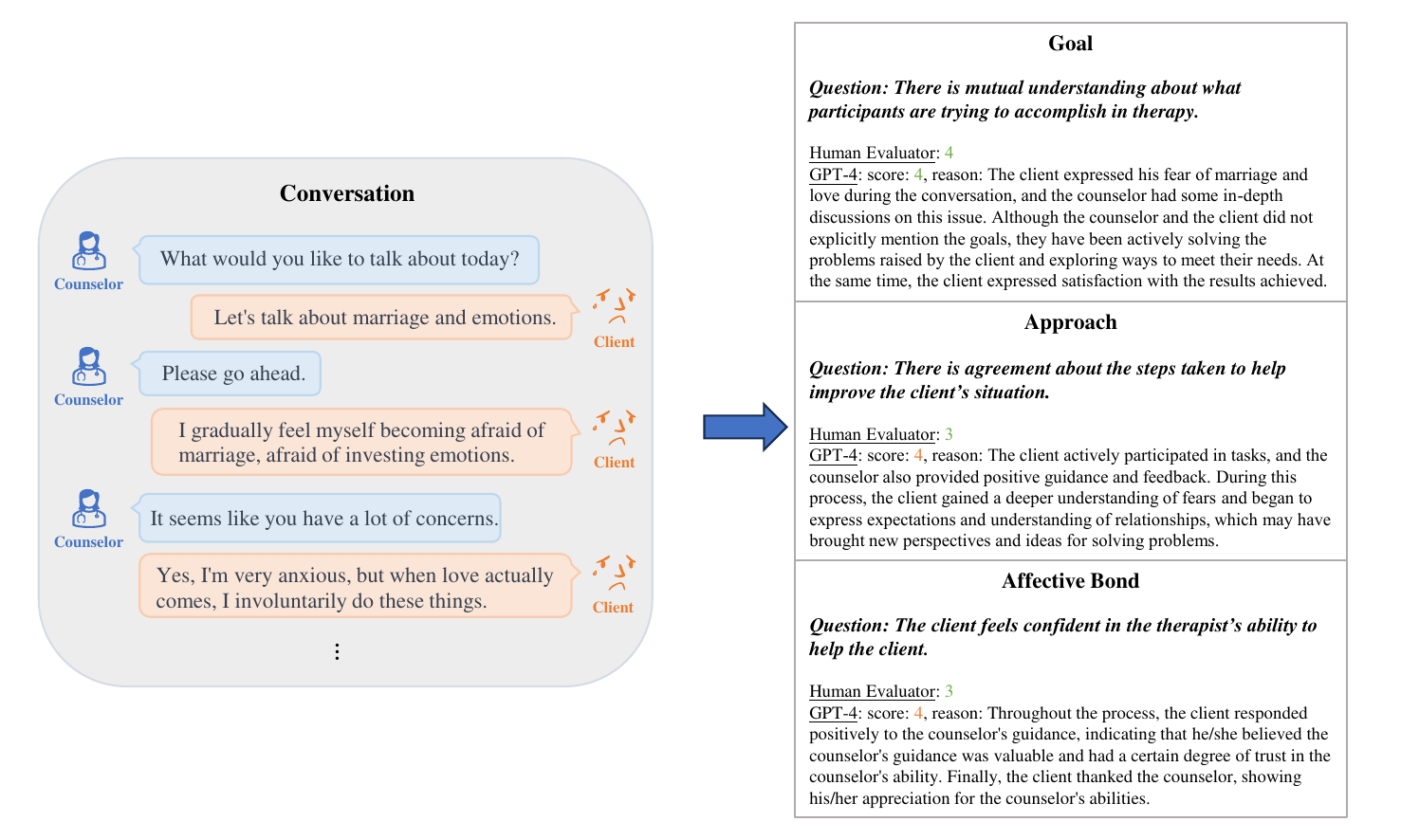}}
    \caption{Our therapeutic alliance framework comprises three integral components: consensus on goal-setting and approaches, and the cultivation of affective bonds. Each component is measured by four questions, each scored with customized guidelines, distinguishing between substantial evidence against, some evidence against, no evidence against, some evidence for, and substantial evidence for these aspects.}
    \label{fig:intro-fig}
\end{figure*}

In this paper, we present an effective automatic approach using Large Language Models (LLMs) to understand the establishment of therapeutic alliances in online text-based counseling~\citep{wei2022emergent}. We propose a conceptual framework with detailed guidelines to characterize the alliance in conversations, adapting theories and scales of therapeutic relationships from face-to-face therapy to text-only interactions. The framework includes three dimensions: goal-setting consensus, approaches to goals, and cultivation of affective bonds (Figure~\ref{fig:intro-fig}). The guidelines facilitate identifying observed elements in textual conversations corresponding to each framework component.

We then collect a large-scale text-based counseling dataset from an online platform. Using our proposed framework and guidelines, trained experts annotate a subset of sessions with high inter-rater reliability. We employ prompt tuning to enable LLMs to apply these guidelines in understanding the alliance within texts. Additionally, we use the Chain-of-Thought (CoT) process~\citep{wei2022cot} to help models identify supportive evidence for their evaluations (as shown in the \textit{reason} part of Figure~\ref{fig:intro-fig}). Experimental findings show that integrating precise guidelines and CoT significantly enhances LLMs’ ability to understand the alliance, ensuring consistency and alignment with experts.

We use the best-performing model on the remaining unannotated sessions to show a positive correlation between the alliance and favorable counseling outcomes. Our findings highlight that counselors, including experienced ones, may struggle to build deeper connections as counseling progresses. This underscores the need for evaluation and feedback mechanisms to enhance counseling effectiveness. Our small-scale proof-of-concept demonstrates that LLM-based feedback can offer counselors insights to better understand their alliances with clients and improve their relationship-building skills.

\section{Related Work}

\paragraph{Automatic Evaluation of Counseling Using NLP.}

Many researchers have endeavored to leverage machine learning and NLP techniques for the automatic evaluation of conversations in mental health counseling, including assessing counselors' therapeutic skills~\citep{cao2019observing, gibson2016deep, chiu2024computational} and treatment fidelity~\citep{atkins2014scaling}, as well as clients' responses to interventions~\citep{tanana2015recursive, li-etal-2023-understanding}. These efforts have predominantly focused on analyzing individual participant behaviors and linguistic features rather than the relational dynamics between counselors and clients. However, in psychotherapy research, the relationship between counselors and clients is extensively studied. The working alliance, defined as the collaboration and attachment between counselors and clients, stands out as a critical researched variable~\citep{bordin1979working_alliance, norcross2010therapeutic, falkenstrom2014working}. Although methods exist for evaluating therapeutic relationships in traditional face-to-face therapy settings~\citep{goldberg2020ml_wa, martinez2019personae_wa, lin2022transformer_wai, tsakalidis2021automatic}, resources tailored to text-based counseling conversations remain scarce. Moreover, these studies often focus solely on specific linguistic features in counselors or clients' utterances or their turn-level interactions, limiting the interpretability of how relationships are established throughout the entirety of the conversation.

Our research is designed to leverage the understanding and reasoning capabilities of LLMs to comprehensively explore the development of critical components of therapeutic alliance through text-only interactions.

\paragraph{LLMs for Mental Health Analysis.}
As the emergence of LLMs showcasing advanced text understanding and reasoning capabilities, recent research has explored to leverage LLMs in mental health analysis~\citep{ji2023rethinking, demszky2023using}. Most studies focuses on analyzing users' posts published in social media platforms to predict their personality traits~\citep{amin2023affective}, sentiment~\citep{zhang2023sentiment}, and mental health conditions~\citep{xu2024mentalLLM, amin2023affective, yang2023interpretable, lamichhane2023evaluation}, including anxiety, depression, suicide ideation, and others. Besides, several studies investigate multi-turn mental health counseling conversations to improve efficiency in psychological counseling~\citep{adhikary2024exploring, han2024chain, lee-etal-2024-towards}. ~\citet{adhikary2024exploring} utilize LLMs to summarize counseling sessions across specific components such as patients' symptoms and history, patient discovery, and reflection, aiding counselors in treatment planning. ~\citet{han2024chain} proposes a Chain-of-Interaction prompting method to empower LLMs in identifying client behavior during motivational interview counseling sessions. ~\citet{lee-etal-2024-towards} employ GPT models to comprehend crisis counseling dynamics, including counselor techniques at the utterance level and client-reported counseling outcomes at the session level. 

Different from the existing studies, our work aims to empower LLMs to better understand interlocutors' relationships in the context of text-based counseling, which is a critical process variable in counseling. This task presents a greater challenge for LLMs, as it demands a heightened capability to comprehend natural language and human interactions within the mental health domain.

\section{Framework and Guidelines for Measuring Therapeutic Alliance}
To understand the alliance between counselors and clients in psychological counseling, we adapt the existing therapeutic alliance definitions and scales to the context of online text-only counseling. To facilitate accurate understanding based on this framework, we carefully design specific guidelines in collaboration with counseling psychology experts.

\subsection{Framework}
In psychology research, the preeminent definition of therapeutic alliance, as introduced by~\citet{bordin1979working_alliance}, emphasizes interactive and collaborative elements in counselor-client relationship in the context of a positive affective attachment~\citep{constantino2002working}. This concept consists of three core components -- counselors and clients' mutual agreement on the targets of counseling (\textit{Goal}), abilities to engage in the tasks of counseling (\textit{Approach}), as well as the cultivation of emotional connections (\textit{Affective Bond})~\citep{bordin1979working_alliance}.

We adopt the Observer-rated Short version of Working Alliance Inventory (WAI-O-S)~\citep{tichenor1989comparison_wa} to measure the alliance. This inventory comprises 12 designed questions, with each alliance dimension measured by four questions. Each question is rated ranging from 1 to 5 points. Its reliability and validity has undergone thorough and comprehensive verification in various psychotherapy types~\citep{santirso2018validation, ribeiro2021observing}. Table~\ref{tab:wai} presents the dimensions along with questions that shape the alliance.

\begin{table*}[]
\centering
\scalebox{0.85}{
\begin{tabular}{|m{3cm}<{\centering}|p{13cm}|m{1cm}<{\centering}|}

\hline
\textbf{Dimension}    & \makecell[c]{\textbf{Question}} & \textbf{No.}                                                                                                  \\ \hline
\multirow{4}{*}{Goal} & There is mutual understanding about what participants are trying to accomplish in therapy.    & Q1                                \\ \cline{2-3}
                      & The client and counselor are working on mutually agreed upon goals.                                & Q2                                          \\ \cline{2-3}
                      & The client and counselor have same ideas about what the client’s real problems are.                 & Q3                                    \\ \cline{2-3}
                      & The client and counselor have established a good understanding of the changes that would be good for the client.      & Q4                       \\ \hline
\multirow{4}{*}{Approach} & There is agreement about the steps taken to help improve the client’s situation.                                          & Q5                   \\ \cline{2-3}
                      & There is agreement about the usefulness of the current activity in therapy (i.e., the client is seeing new ways to look at his/her problem). & Q6 \\ \cline{2-3}
                      & There is agreement on what is important for the client to work on.      & Q7                                                                     \\ \cline{2-3}
                      & The client believes that the way they are working with his/her problem is correct.    & Q8                                                        \\ \hline
\multirow{4}{*}{Affective Bond} & There is a mutual liking between the client and counselor.                  & Q9                                                                 \\ \cline{2-3}
                      & The client feels confident in the counselor’s ability to help the client.        & Q10                                                            \\ \cline{2-3}
                      & The client feels that the counselor appreciates him/her as a person.          & Q11                                                               \\ \cline{2-3}
                      & There is mutual trust between the client and counselor.            & Q12                                                                          \\ \cline{2-3}
                      \hline
\end{tabular}
}
\caption{The framework of working alliance contains three core components: \textit{Goal}, \textit{Approach}, and \textit{Affective Bond}. Each dimension is assessed through a set of four questions.}
\label{tab:wai}
\end{table*}

\paragraph{Goal.}
In counseling, goals are important for facilitating changes in clients' thoughts, feelings, and actions. They provide direction for both counselors and clients during their sessions. Clear agreement on goals increases adherence and leads to better outcomes. However, at the beginning of counseling, there can be a lack of clarity about clients' issues and differences in goals between clients and counselors. To address this, counselors should engage in deeper discussions with clients to establish mutually endorsed and valued objectives.

\paragraph{Approach.}
In addition to the agreement on goals, the strength of the working alliance also depends on the participants' clear and mutual understanding as well as acceptance on the tasks that their shared goals impose upon them~\citep{bordin1983wa_supervision}. Tasks are usually assigned by counselors based on their counseling styles, personal experiences and predispositions. However, clients may not fully understand the interconnections between the assigned tasks and the overarching goals. Moreover, clients may perceive that the demands of tasks exceed their abilities. In such cases, counselors need to skillfully adapt to their clients by offering alternative or modified tasks, thereby empowering clients to actively and effectively engage.

\paragraph{Affective Bond.}
Apart from cognitive collaboration, emotional connections play a crucial role in shaping the therapeutic alliance. The concept of affective bonds embraces the complex network of positive personal attachments between counselors and clients, including issues such as mutual trust, liking, acceptance, and confidence~\citep{horvath1990development}. As clients perceive that counselors genuinely care about and appreciate them, a sense of security is established, fostering a greater willingness to delve into deeper self-disclosure during counseling, particularly in discussing their negative behaviors and thoughts. Moreover, clients' confidence in counselors' capabilities to facilitate positive changes make them more inclined to accept counselors' guidance and actively participate in the tasks assigned by the counselors.

\subsection{Guidelines}
To facilitate the understanding of questions and the differentiation of scores in text-only exchanges, we have four developers to carefully design specific guidelines to each question-score combination. 

We formulate our guidelines based on two main principles guided by psychology theories on scale and guideline development~\citep{mahalik1994developmentCRS, form2000manual}: (1) determining observable elements within textual conversations to reflect subjective relationship-related questions, and (2) ensuring a balanced score scale that assumes an average rather than a positive alliance between counselors and clients, thus mitigating potential ceiling effects. Concretely, we derive behavioral or attitudinal indicators from literature~\citep{bordin1979working_alliance, doran2016working, form2000manual}. Subsequently, we outline the frequency of behaviors and intensity of attitudes at each score level, with a neutral point set at 3 as the start point. We then task our developers with iteratively refining the guidelines through application to counseling sessions. After three iterations of repeating annotation on 15 conversations, we finalized the guidelines. The intra-class agreement ICC~\citep{koo2016icc_guideline} among the four developers in the three iterations are as follows: 0.5267, 0.6084, and 0.6603. The monotonically increasing agreement proves that the iterative process effectively resolves differences among developers. And the moderate agreement ensures the reliability of our guidelines. More details on the developers, development process and the finalized guidelines are presented in Appendix~\ref{framework_development}.

\section{Data Collection}
To validate the feasibility of our proposed framework, we collect counseling conversations between professional counselors and actual clients, and carefully annotate a subset of these conversations according to the framework.

\subsection{Data Source}

We developed an online text-based counseling platform and enlisted 9 qualified professional counselors (7 females; \textit{Age range}: 25 $\sim$ 45 years old, \textit{Mean} = 34.67, \textit{SD} = 7.45). We also recruited 82 adults (55 females; \textit{Age range}: 19 $\sim$ 54 years old, \textit{Mean} = 27.62, \textit{SD} = 5.94) as clients who were voluntarily and eligible for online psycho-counseling. To be responsible for our participants, all these clients were assessed using the self-report symptom inventory (SCL-90)\cite{wxd1999} to ensure they did not exhibit severe depressive, anxious, or psychiatric symptoms. Each client was assigned to a counselor, with the number of clients assigned to each counselor ranging from 4 to 13. Each counseling session lasted 50 minutes, which is a widely accepted standard duration for psychological counseling. Clients were encouraged to attend a minimum of 7 counseling sessions, scheduled weekly or bi-weekly.

We collected total 859 counseling sessions. The statistics of the overall conversations are detailed in Table~\ref{tab:data_statistics}. The length of counseling conversations are significantly longer than the existing conversations obtained through crowdsourcing or generated by language models (avg. 76.07 utterances compared to 29.8 utterances in ESConv~\citep{liu2021towards} and 6.36 utterances in SMILE~\citep{qiu2023smile}). Moreover, each counselor-client pair engages in multiple consecutive counseling sessions (avg. 10.48 sessions compared to 4 sessions in Multi-Session Chat~\citep{xu-etal-2022-goldfish}), suggesting, in real-world scenarios, an effective resolution of clients' concerns often requires extended multi-turn interactions and multiple sessions.

\begin{table}[h]
\centering
\scalebox{0.75}{
\begin{tabular}{@{}cccc@{}}
\toprule
\textbf{Category}                     & \textbf{Total} & \textbf{Counselor} & \textbf{Client} \\ \midrule
\textbf{\# Dialogues}                 & 859          & -                  & -               \\
\textbf{\# Speakers}                  & 91          &  9              & 82             \\
\textbf{\# Avg. sessions per speaker} & -          &  95.44               & 10.48              \\
\textbf{\# Utterances}                & 65,347         &     32,860         &  32,487         \\
\textbf{Avg. utterances per dialogue} & 76.07    &   38.25           &  37.82         \\
\textbf{Avg. length per utterance}    & 26.84    &  24.01            &   29.70        \\ \bottomrule
\end{tabular}}
\caption{Statistics of the overall conversations.}
\label{tab:data_statistics}
\end{table}

\subsection{Annotation Process}
To ensure the quality of the annotations, we engaged 3 experienced developers of the guidelines to annotate a subset of collected conversations. Their extensive knowledge of the working alliance framework and guidelines allowed for a thorough evaluation. Before the annotation process, we took measures to protect the privacy of the counselors and clients by anonymizing their personal information, including names, organizations, addresses, and more.

For the annotation phase, we randomly selected 79 sessions involving 4 counselors and 8 clients. Each conversation was annotated by all three annotators. To determine the final score for each question, we calculated the average of all scores assigned by the annotators.

After obtaining the annotated data, we calculated the intraclass correlation coefficient (ICC)\citep{koo2016icc_guideline} among the annotators for each question. The inter-rater agreement for the dimensions of \textit{Goal}, \textit{Approach}, and \textit{Affective Bond} were found to be 0.7581, 0.6587, and 0.6498, respectively. These values indicate a reliable level of agreement among the annotators\footnote{An ICC value between 0.5 and 0.75 indicates moderate reliability, while a value between 0.75 and 0.9 indicates good reliability.}. Further details regarding the inter-rater agreement for each question can be found in Appendix~\ref{appendix: human_agreement}.

\subsection{Data Characteristics}

\begin{figure}
    \centering
    \scalebox{0.35}{
    \includegraphics{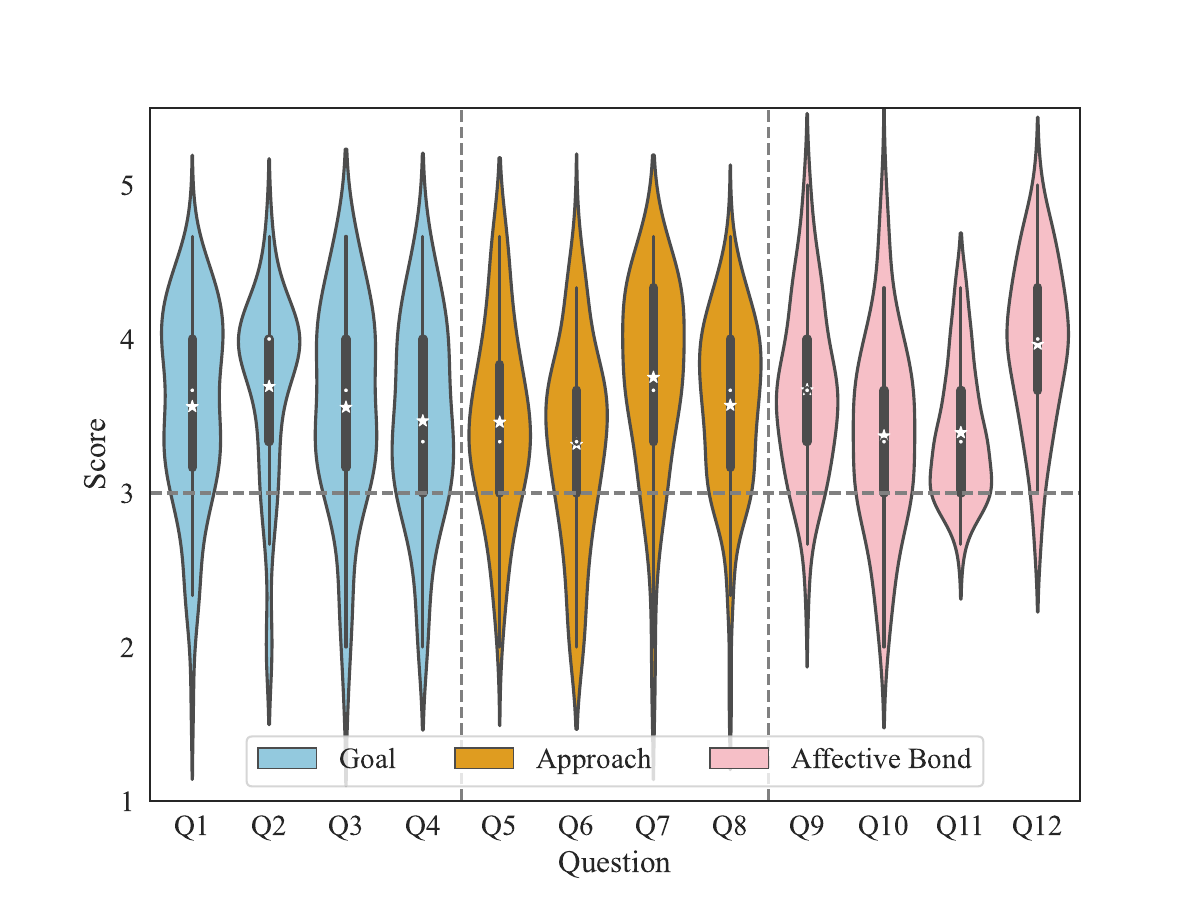}}
    \caption{The violin plot of the distribution of scores annotated for each question, with a boxplot inside. The white pentagons within the violins represent the mean values.}
    \label{fig:score_distribution}
\end{figure}

Figure~\ref{fig:score_distribution} illustrates the distribution of annotated scores for all the questions. For further insights into the average scores per dimension and question, as well as their corresponding standard deviations, please refer to Appendix~\ref{appendix: data_avg_scores}.

On average, the scores for each dimension range between 3.5 and 4, indicating that counselors are able to establish relatively positive relationships with clients through text-based communication, yet there remains significant room for improvement. Among the three dimensions of alliance, the \textit{Affective Bond} stands out with the highest average score, particularly in the question regarding mutual trust between counselors and clients (Q12), where the score almost reaches 4. This suggests that a strong sense of trust can indeed be cultivated, enabling clients to openly discuss personal concerns. However, the \textit{Goal} and \textit{Approach} dimensions have the relatively lower average scores, specifically in the question concerning agreement on the usefulness of the current therapy activity (Q6, avg. = 3.32). This signifies the need for clearer counseling goals and connecting therapeutic activities to these goals to enhance client engagement.

\section{LLM Evaluation}
With annotated data and proposed guidelines, we conduct zero-shot experiments to prompt advanced LLMs including GLM-4, Claude-3, ChatGPT and GPT-4 to understand the therapeutic relationships reflected in text-only conversations.

\subsection{Setup}

The prompt comprises four key components: the definition of evaluation task, the counseling conversation to be evaluated, the evaluation question and corresponding guidelines. To further investigate the impact of guidelines on the evaluation capabilities of LLMs, we conduct three experimental settings — prompting LLMs without guidelines, with general guidelines, and with our proposed detailed guidelines. Under the general guidelines, each question is accompanied with the uniform criteria: substantial evidence against, some evidence against, absence of evidence for or against, some evidence for, and substantial evidence for the item. The impact of CoT process on the scoring of LLMs after providing detailed evaluation criteria is also explored. In the CoT setting, we require models to provide corresponding evidence for ratings within the dialogue text. We carefully design specific prompts for each experiment setting accordingly. Example prompts are illustrated in Figure~\ref{fig:prompt} in Appendix~\ref{appendix:llm_evaluation_results}.

\subsection{Models}
We select four accessible top-performing large language models -- GLM-4 (Zhipu AI)~\citep{ZHIPU2024GLM-4}, Claude-3 (\textit{Sonnet} model; Anthropic)~\citep{Anthropic2024}, ChatGPT (\textit{gpt-35-turbo-16k} model; OpenAI)~\citep{chatgpt} and GPT-4 (\textit{gpt-4} model; OpenAI)~\citep{2023gpt4}. These models have been enhanced to follow human instructions through instruction tuning and align with human preferences via reinforcement learning from human feedback (RLHF, ~\citep{Ouyang2022RLHF}). Our interactions with these models are facilitated using the official API. The temperature and nuclear sampling parameter are set as 1.0 for all models. Each model is tasked with rating the same conversation three times independently for thorough evaluation.

\subsection{Results and Analysis}

\paragraph{Model Self-Consistency.}

The reliability of a model as an annotator depends on its consistency in multiple independent evaluations of the same samples. We evaluate all these models by assessing their consistency across all the experimental settings. The results of models' self-consistency are shown in Table~\ref{tab:alignment_results}, and detailed results can be found in Table~\ref{tab:model_agreement} in the Appendix.

We find that ChatGPT falls short of reaching a moderate level of self-agreement without detailed guidelines and CoT. However, GLM-4, Claude-3, and GPT-4 maintain a moderate or higher level of self-consistency, ensuring the validity of their annotated results. Therefore, we further analyze the influence of guidelines and CoT on the alignment between these latter three models and human evaluations in the following.

\begin{table*}[]
\centering
\scalebox{0.78}{
\begin{tabular}{ccccccc}
\hline
\multicolumn{2}{c}{\textbf{Models}}                                     & \textbf{ICC} & \textbf{Goal}            & \textbf{Approach}        & \textbf{Affective Bond}  & \textbf{Overall}                 \\ \hline
\textbf{ChatGPT}                   & \textbf{Detailed Guidelines + CoT} & 0.5209       & 0.2004                   & \textit{0.3612}          & \textit{0.4122}          & \textit{0.3246} \\ \hline
\multirow{4}{*}{\textbf{GLM-4}}    & \textbf{No Guidelines}             & 0.9955       & 0.3187                   & 0.4117                   & 0.4466                   & 0.3924                           \\
                                   & \textbf{General Guidelines}        & 0.9921       & 0.3723                   & 0.4844                   & 0.4300                   & 0.4289                           \\
                                   & \textbf{Detailed Guidelines}       & 0.9960       & \textit{0.4184}          & 0.4301                   & 0.4893                   & 0.4459                           \\
                                   & \textbf{Detailed Guidelines + CoT} & 0.9938       & 0.4102                   & \textit{0.5004}          & \textit{0.4997}          & \textit{0.4701}                  \\ \hline
\multirow{4}{*}{\textbf{Claude-3}} & \textbf{No Guidelines}             & 0.7408       & 0.3821                   & 0.4713                   & 0.3506                   & 0.4013                           \\
                                   & \textbf{General Guidelines}        & 0.8240       & 0.3229                   & 0.4724                   & 0.3962                   & 0.3971                           \\
                                   & \textbf{Detailed Guidelines}       & 0.7823       & \textit{0.4700}          & 0.4506                   & \textit{\textbf{0.5024}} & 0.4743                           \\
                                   & \textbf{Detailed Guidelines + CoT} & 0.8322       & 0.4552                   & \textit{\textbf{0.5608}} & 0.4787                   & \textit{0.4982}                  \\ \hline
\multirow{4}{*}{\textbf{GPT-4}}    & \textbf{No Guidelines}             & 0.6687       & 0.3591                   & 0.4288                   & 0.3693                   & 0.3857                           \\
                                   & \textbf{General Guidelines}        & 0.7482       & 0.3320                   & 0.4516                   & 0.3961                   & 0.3933                           \\
                                   & \textbf{Detailed Guidelines}       & 0.6854       & \textit{\textbf{0.4979}} & \textit{0.5480}          & 0.4417                   & 0.4959                           \\
                                   & \textbf{Detailed Guidelines + CoT} & 0.7205       & 0.4937                   & 0.5448                   & \textit{0.4667}          & \textit{\textbf{0.5018}}         \\ \hline
\end{tabular}}
\caption{The inter-rater reliability and overall Pearson correlation results of all models with human evaluation on the working alliance dimensions across different experimental settings. The \textit{italics} highlight the best results obtained by each model under different experimental settings, whereas the \textbf{bold} emphasizes the best results achieved across all models and experimental configurations. The overall column presents the average scores across the goal, approach, and affective bond dimensions here.}
\label{tab:alignment_results}
\end{table*}

\paragraph{Alignment with Human Evaluations.}
The models' capability on understanding the working alliance is defined as the extent to which its assessments align with those of human experts. Table~\ref{tab:alignment_results} summarizes the Pearson correlation coefficients~\citep{lee1988correlation_coefficient} between LLMs and human evaluations across different experimental settings. Results indicate that GPT-4, accompanied by detailed guidelines and CoT, exhibits the best average  performance over the three aspects compared to alternative models and experimental setups.

\indent\textbf{Guidelines.} As shown in Table~\ref{tab:alignment_results}, the overall results indicate a trend wherein an increase in the level of detail in guidelines enhances alignment. This improvement is particularly significant when transitioning from general guidelines to more detailed ones, resulting in a notable average increase in correlation of 23.61\%. Detailed guidelines are particularly effective in enhancing LLMs' performance on challenging questions. For instance, in the case of discerning whether counselors and clients like each other (Q9), GPT-4 performs poorly without guidelines or with general guidelines. However, when detailed guidelines are provided, there is a remarkable 76\% increase in correlation (Detailed results can be found in Table \ref{tab:pearson_correlation_5_likert} in the Appendix). 

These findings highlight the potential to improve the alignment of LLM evaluations with human assessments by refining the guidelines. Simultaneously, we emphasize the significance of self-consistency in LLMs, which is a crucial prerequisite for their effectiveness as evaluators.

\indent\textbf{Chain-of-Thought Prompting.} Table~\ref{tab:alignment_results} demonstrates that integrating CoT improves the alignment of LLM evaluations with human assessments across overall dimensions. CoT significantly enhances LLMs' performance on challenging questions. For instance, with regard to the challenging question Q9 mentioned above for GPT-4, incorporating CoT leads to a significant 32.05\% increase in the Pearson correlation with human evaluations. Thus, facilitating evidence extraction and explanation generation prior to scoring proves to be an effective strategy for enhancing LLMs' comprehension of dialogue content and improving assessment accuracy.

\section{LLM-based Insights into Text-based Mental Health Counseling}
We employ the best-performing model (i.e., GPT-4 with detailed guidelines and CoT) to study how the therapeutic alliance impacts online text-based psychological counseling. We utilize the model to predict the alliance in the remaining unannotated sessions. We investigate how counselors' experience and counseling progress influence the alliance strength, and examine its correlation with counseling outcomes. Furthermore, we analyze common patterns in counselors' interactions that affect the establishment of the therapeutic relationship. Additionally, we showcase the efficacy of LLM-based feedback for counselors through a simple proof-of-concept demonstration.

\paragraph{Counselors' Counseling Experience $\neq$ Abilities to Establish Relationships.}

We explore whether counselors with more experience find it easier to establish therapeutic relationships with clients. Nine counselors are grouped by their counseling experience: primary (≤ 2 years), intermediate (3-8 years), and advanced (≥ 10 years). We calculate average working alliance scores for each counselor across all sessions to gauge their relationship-building proficiency, with t-tests~\cite{kim2015ttest} to reveal their potential differences.

Results (Figure~\ref{fig:counselor_avg_score} and Figure~\ref{fig:ttest_counselor_score} in Appendix~\ref{appendix:counselor_ability}) show that extensive experience does not necessarily correlate with stronger therapeutic connections. Counselor I, at the intermediate level, excels in goal-setting and emotional rapport, outperforming advanced peers. Conversely, Counselor E, also intermediate, falls behind even novices, especially in goal-oriented approaches. These observations align with previous research suggesting counselors may experience skill regression over time without intentional practice and constructive feedback~\citep{goldberg2016psychotherapists, sharma-etal-2020-computational-empathy}, emphasizing the need for continuous evaluation and feedback.

\begin{figure}
    \centering
    \scalebox{0.28}{
    \includegraphics{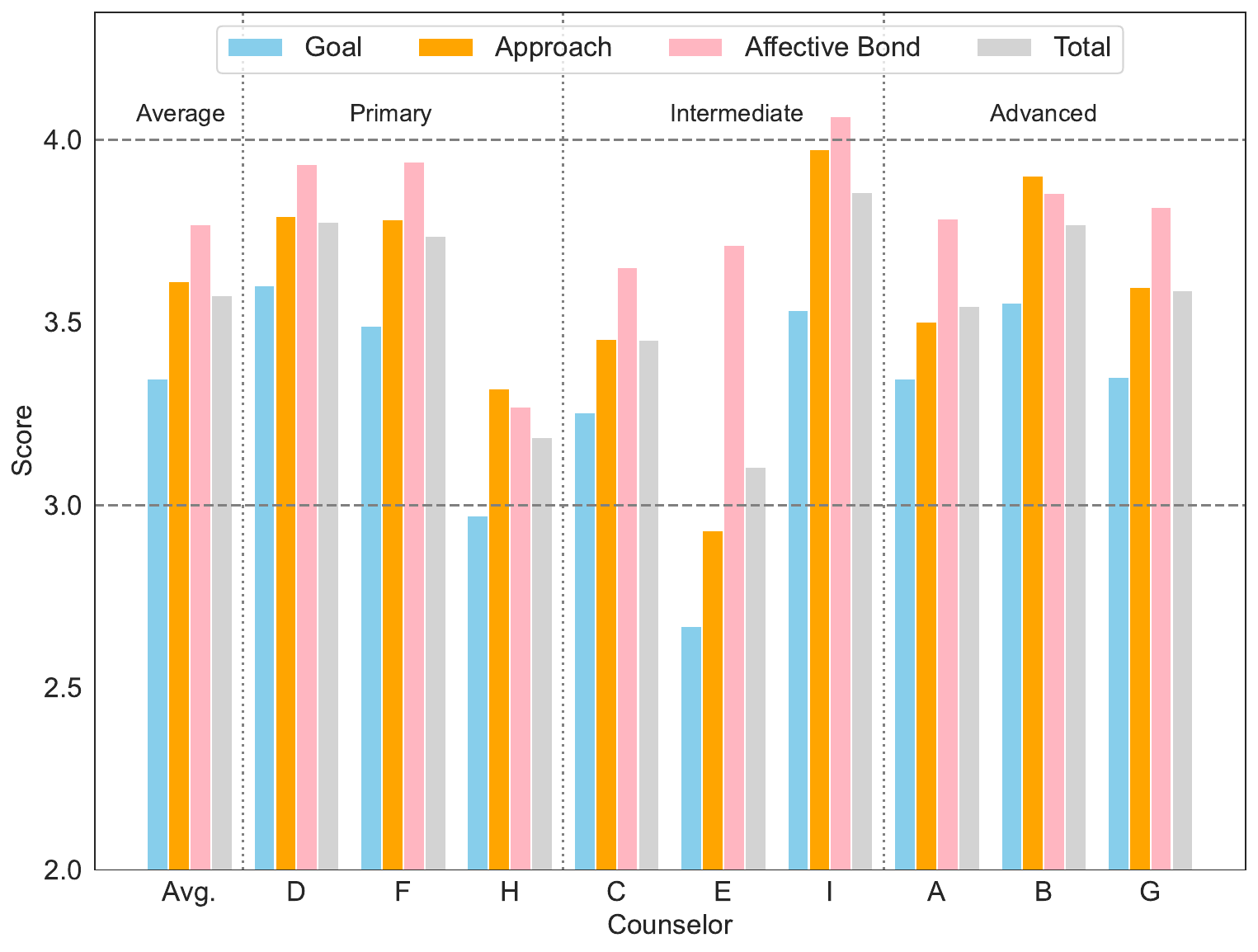}}
    \caption{The average alliance scores for all counselors and counselors with varied experience levels.}
    \label{fig:counselor_avg_score}
\end{figure}

\paragraph{Long-Term Communications $\neq$ Stronger Alliance.}

We divide clients’ counseling sessions into three phases—early, middle, and late—and compare the counselor-client relationship across these stages. Our findings indicate that the relationship does not significantly deepen over time. Specifically, there is only a marginal increase in affective connections, while agreement on counseling goals and approaches remains constant. Further analysis shows that nearly 50\% of client-counselor pairs experience either a decline or no change in the strength of the therapeutic alliance, with less than 3\% improving by at least one level within our framework. This emphasize the challenges counselors face in enhancing relationship-building skills without adequate feedback.

\paragraph{Better Counseling Outcomes are More Likely Based on Robust Alliance.}

Psychology research underscores the pivotal role of a robust alliance in counseling outcomes~\citep{horvath1994wa, falkenstrom2014working}. Here, we utilize clients' self-reported ratings on the Outcome Rating Scale (ORS)~\citep{miller2003outcome, bringhurst2006reliability} to gauge the effectiveness of each counseling session. The ORS evaluates various aspects including clients' individual physical and mental well-being, interpersonal relationships, social role functioning and overall well-being, with scores ranging from 0 to 100 for each aspect. Pearson correlation analyses between total working alliance scores and ORS dimensions show significant correlations ($r \approx 0.30$, $p$ < 0.001). This indicates that a stronger working alliance may be associated with more favorable conversation outcomes. Additional details are provided in Appendix~\ref{appendix: wa_outcomes}.

\begin{figure}
    \centering
    \scalebox{0.5}{
    \includegraphics{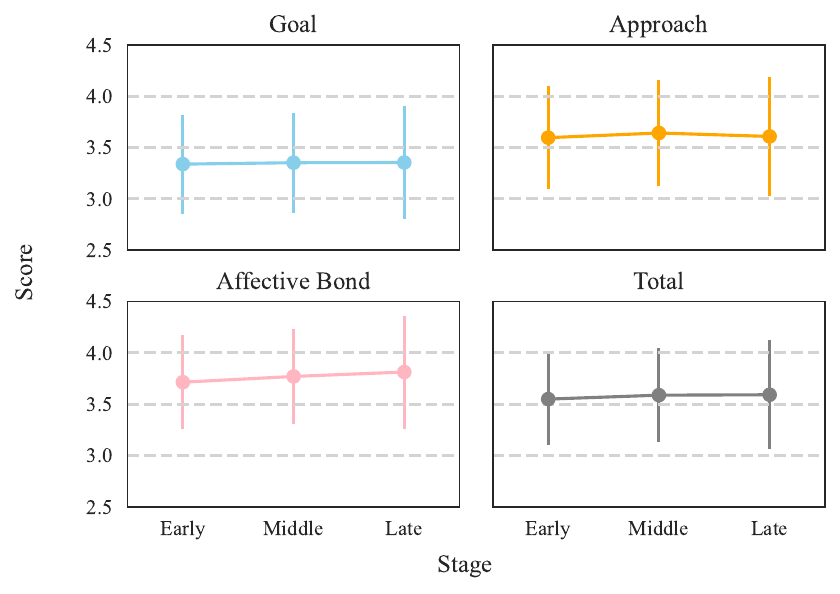}}
    \caption{The average alliance across counseling stages.}
    \label{fig:wa_over_time}
\end{figure}

\paragraph{Counselors' Common Behaviors Shape the Therapeutic Alliance.} We conduct further natural language analysis to investigate common counselor behaviors that influence the establishment of therapeutic relationships, based on GPT-4-generated explanations.

By analyzing and comparing counselors' performance in sessions with both poor and strong alliances, we have gained valuable insights. Counselors who struggle to establish a robust alliance often overlook their clients' negative emotions and resistance during counseling. Their guiding approach typically manifests in two problematic ways: either they respond passively to clients' statements without exploring core issues, or they overstep boundaries by excessively directing clients, thus compromising their autonomy. Additionally, the feedback provided tends to be vague and generalized, lacking the personalized and specific solutions that address each client's unique needs.

These findings highlight the necessity for counselors to cultivate deeper empathy and consistently monitor clients' responses and behaviors throughout the counseling process. Counselors must find a balance between offering guidance and respecting clients' autonomy. It is crucial for them to provide tailored advice that deeply considers each client's circumstances to foster a positive therapeutic alliance.

\paragraph{Implications for Feedback using LLMs' Explanations.}

These results highlight the significance of the working alliance in online text-based counseling, supported by clients' positive outcomes and psychological theories. However, even experienced counselors may face challenges in fostering deep connections in long-term sessions. To address this, we use GPT-4' explanations generated via CoT to provide constructive feedback to counselors.

Counselors E and H, who struggled with relationship-building, were given 10 sessions evaluated by LLM with explanations as feedback. They assessed the feedback on: (1) enhancing their understanding of the client alliance, (2) identifying improvement directions, and (3) willingness to adjust strategies based on the feedback. Each aspect was rated from 1 to 5, with 3 as neutral. The average scores were 3.43, 3.49, and 3.74, respectively, indicating the efficacy of LLM-based feedback in helping counselors deepen their client alliances and refine their connection-building approach. Further details are in Appendix~\ref{appendix:llm-based_feedback}.

In future work, we will integrate LLM-based real-time evaluation and feedback on the working alliance into actual counseling sessions to facilitate counselors cultivate deeper therapeutic connections with their clients.

\section{Conclusion}

We adapted a framework with crafted guidelines, a dataset, and LLM-based approaches to understand the working alliance between counselors and clients in online text-based counseling. Our findings indicate that integrating detailed guidelines and CoT prompting enables LLMs to assess the working alliance with underlying rationales effectively. Furthermore, we underscore the importance of the working alliance in online counseling and demonstrate the utility of LLM-based feedback in enhancing counselors’ understanding of their client relationships and providing valuable insights for improvement.

\section{Limitations}

As the first LLM-based approach to automatically understand the working alliance in online text-based counseling, there is significant potential for future improvement. This paper focuses on using prompt tuning to leverage LLMs for evaluating therapeutic relationships between counselors and clients. We iteratively refine guidelines with human involvement to enhance LLM capabilities in assessing the working alliance. While effective, this approach is time-consuming and labor-intensive. Automating guideline optimization through model self-improvement is a promising alternative to streamline the process. Additionally, fine-tuning techniques can further enhance model performance once sufficient paired data of counseling dialogues and alliance scores are available. 

It is also important to emphasize that our analysis and findings should be interpreted with caution. First, our dataset is limited in both size and diversity, with only 9 counselors included. Second, participants were aware that their (anonymized) data might be shared, which could have influenced their interactions, as volunteers may have been hesitant to fully disclose due to privacy concerns.

Our preliminary proof-of-concept validation indicates that LLM-based feedback helps counselors better understand their relationships with clients and provides guidance for improving alliance-building skills. Further research is needed to assess the efficacy of providing counselors with prompt LLM-based feedback after each session in real online counseling settings, aiding in fostering stronger client relationships and enhancing counseling effectiveness. We leave this as a task for future research.

\section{Ethics Statement}

\paragraph{Data Privacy.} This study is granted ethics approval from the Institutional Ethics Committee. All counselors and clients consented to participate and receive reasonable fee for participation. Participants were informed that the conversations collected on the platform would be utilized for scientific research purposes and potentially shared with third parties for this purpose. Participants were also informed that they could discontinue counseling and withdraw from the research at any time. The detailed consent form for clients and user services agreement are presented in Appendix~\ref{consent form}.

Throughout the annotation process, we devoted meticulous attention to manually de-identifying and anonymizing the data, ensuring the utmost protection of the privacy of both clients and counselors. Additionally, our guidelines developers and annotators, prior to accessing the conversation data, formally committed to data confidentiality agreements and adhered to ethical guidelines, underscoring our commitment to upholding the highest standards of privacy and ethical conduct. Moreover, to avoid potential privacy concerns during LLMs evaluations, we utilize LLMs through the official API and provide them with the anonymized data. 

\paragraph{Data Release.} In order to foster interdisciplinary research at the intersection of NLP and psychology, we plan to release a subset of this dataset to interested researchers upon article acceptance. For whom request the data, we will evaluate their qualification. We require them to provide a valid ID, the reason they request data, proof of full-time work in non-profit academic or research institutions which have the approval of an Institutional Review Board (IRB), full-time principal investigators, and the approval of the institution’s Office of Research or equivalent office. Meanwhile, they must sign a Data Non-disclosure Agreement and promise that they would not share the data with any third party.

\paragraph{LLM-based Feedback.}  We advocate for the utilization of LLM-based feedback as a auxiliary and guiding tool for counselors to discern shortcomings in counseling sessions and offer pathways for potential enhancement, rather than replacing expert evaluations. We assert that when employing LLM-based feedback, it's imperative to consider the following issues:

\textit{1) Imperfect Capabilities of LLMs:} Due to current limitations, LLMs may not perfectly align with assessments by professional evaluators. Relying on inaccurate results from LLMs could undermine the effectiveness of psychological counseling and create medical and legal risks.

\textit{2) The Opacity of LLMs:} Given LLMs' inherent lack of transparency, it is crucial to exercise great caution when interpreting the explanations and evaluations obtained from them.

\textit{3) Societal Acceptance:} There is uncertainty about the societal acceptance of LLM-based feedback for counselors. Counselors with lower acceptance of AI may be hesitant to accept feedback from LLMs. Concerns regarding potential technology misuse and ethical issues related to human-machine collaboration may lead to public resistance and opposition to the application of LLMs.

\section*{Acknowledgement}
This work is supported by the Key Research and Development Program of Zhejiang Province of China (Grant No.2021-C03139), and the Westlake University Research Center for Industries of the Future (Grant No.WU2023C017).

\bibliography{custom}

\appendix
\section{Guidelines}
\label{framework_development}
\subsection{Guideline Developers}
To ensure the quality in our guideline design, we collaborate closely with psychology experts. We have four developers to carefully design guidelines. One is a postdoctoral fellow in experimental and counseling psychology, holding a State-Certificated Class 3 Psycho-counselor designation with 4 years of practical experience. Another developer holds a master's degree in applied psychology. The remaining two developers specialize in NLP research, with a keen focus on its application in psychological counseling. These two developers have received training in therapeutic techniques and crisis intervention, conducted extensive literature reviews, and possess a comprehensive understanding of counseling practices. Moreover, these developers have collaborated on projects at the intersection of NLP and psychological counseling, demonstrating their extensive experience in the field. Thus, the development of the guidelines are well balanced based on professional views of psycho-counselling and application of NLP techniques.

\subsection{Guidelines Development Process}
\paragraph{Main Process of Guideline Refinement.} Our developers carefully design specific guidelines for each score associated with each question. Following ~\citet{form2000manual}'s work, we employ the amount of evidence present in counseling conversations as anchor labels for scores, using the middle point (i.e., 3) as the start point representing "no evidence". The higher score denotes more positive evidence, and vice versa. As a result, each question is scored from 1 to 5.

Expanding on the general guidelines, we further design specific descriptions for each score of every question. Here, we introduce the detailed descriptions by taking the question "\textit{There is agreement about the usefulness of the current activity in therapy (i.e., the client is seeing new ways to look at his/her problem)}" as an example. Firstly, we anchor the extreme scores of the scale with bipolar adjective relevant to this question, resulting in "open claim of useless" at a rating of 1 and "overt statements of usefulness" at a rating of 5. Secondly, we outline counselors and clients' behavioral indicators at each score level, along with the corresponding extent and frequencies. For the exemplar question, the descriptions are formulated based on clients' frequency (always or sometimes) and attitude (actively or passively) towards participating in tasks proposed by counselors. 

The resulting guidelines establish conceptual boundaries among questions within the same dimension and provide clear distinctions among the points on the scale, allowing raters to discern subtle changes in the working alliance with greater reliability. 

\paragraph{Iterative Refinement.} Firstly, we randomly select 15 conversations and ask all the developers to annotate them independently based on general guidelines. After the annotation, the developers discuss the differences and confusions among their annotations in several conversations until reaching a consensus. During this process, they may refine the guidelines by compiling the behavioral indicators of counselors and clients relevant to each question, with the associated degree and frequency at each score level. The developers repeat annotating these conversations based on modified guidelines. After iterating the above step 3 times, the final version of the guidelines is obtained. The intra-class agreement~\citep{koo2016icc_guideline} among the four developers in the three iterations are as follows: 0.5267, 0.6084, and 0.6603. The monotonically increasing agreement proves that the iterative process effectively resolves differences among developers. And the moderate agreement ensures the reliability of our guidelines.

\subsection{Detailed Guidelines}
\begin{center}
    \textbf{------ Goal ------} 
\end{center}

\noindent\textbf{Q1: There are doubts or a lack of understanding about what participants are trying to accomplish in therapy.}

1 = The counselor or the client explicitly mentions the counseling goals and works around the established objectives, such as understanding information related to the goals and methods to achieve them. The relevance of the dialogue to the goals is evident for both the counselor and the client. They may discuss the goals to acknowledge or comment on the usefulness of the therapeutic process.

2 = The counselor and the client do not explicitly mention the goals but are working towards a common objective. The counselor addresses the client's concerns immediately and adjusts the therapeutic process to meet the client's needs. The client is satisfied with the progress made.

3 = There is no evidence to suggest that the counselor and the client have established consistent counseling goals, or there is an equal level of confusion and understanding regarding the goals.

4 = There is disagreement between the counselor and the client regarding counseling goals. While there may be some communication between both parties, the counselor's specific tasks or interventions may be questioned or resisted by the client. The counseling may need to be paused multiple times to adjust the goals. The client may express overall dissatisfaction with the counseling. At this stage, the counselor may take on an "expert" role, sometimes overlooking the client's opinions or therapeutic ideas, and instances where the counselor guides but the client is not engaged may occur. The client may become less emotionally invested.

5 = The counselor and the client have clearly identified different goals, and there are disagreements in the order of issues and solutions in therapy. This inconsistency may lead the client to express strong dissatisfaction with the overall counseling process and goals, possibly mentioning the reasons for participating in therapy. This could further trigger a negative reaction from the counselor. At this stage, it seems challenging for both parties to find common ground, making the therapeutic process difficult.

\noindent\textbf{Q2: The client and therapist are working on mutually agreed upon goals.}

1 = The shift of topics often occurs abruptly, usually without mutual agreement from both parties. This frequent topic shift may result from one party interrupting or disregarding the other's statements. At this stage, significant conflicts exist between the counselor and the client regarding the appropriateness, definition, and boundaries of the goals, leading to confusion in the rhythm and content of the conversation.

2 = Topics may shift before resolution or conclusion, but the transition typically moves from one relevant topic to another related or less related one. This shift can be initiated by either the counselor or the client. At this stage, both parties may express dissatisfaction with the frequent shift of topics or the overall pace of therapy, but friction is relatively minor and has not escalated into apparent conflict.

3 = There may be some ambiguity or uncertainty between the counselor and the client regarding session goals. The current stage of communication lacks clear evidence that both parties have reached a common understanding or collaboration, but there is also no explicit conflict or disagreement. Further communication and discussion may be necessary to clarify expectations and goals to ensure the effectiveness of therapy.

4 = The counselor and the client have made some progress through discussing relevant topics, but there may still be a small amount of disagreement or areas that need further exploration. At this stage, although both parties generally agree on the current direction and topics of therapy, more communication and consensus may be needed to ensure the achievement of goals.

5 = The counselor and the client have achieved complete agreement on goals through in-depth, targeted discussions, and have had highly productive discussions on multiple related topics. At this stage, both parties almost always reach consensus on the current topic identified by the client as a goal and then smoothly transition to another relevant topic. The overall session and communication are very smooth and efficient.

\noindent\textbf{Q3: The client and therapist have different ideas about what the client’s real problems are.}

1 = The counselor and the client have a very clear and consistent understanding of the client's issues and goals. At this stage, there is a strong consensus on problem resolution, with both parties often identifying the same issues and considering therapy sessions highly effective. This indicates that they have formed a close collaborative relationship in the session.

2 = The counselor and the client have a certain level of consensus on the client's issues and goals. While not fully synchronized like the first category, both parties are making efforts to understand each other and demonstrate open and cooperative attitudes in discussions. This indicates that they are working towards establishing a common therapeutic direction and goals.

3 = In the communication between the counselor and the client regarding the client's issues, there is no clear evidence of agreement or disagreement. In the current interaction, there may be neither a clear consensus nor explicit conflict in opinions and feelings on both sides. Further communication and discussion may be needed to clarify the positions and expectations of both parties.

4 = There is some disagreement between the counselor and the client regarding the client's issues. This disagreement may manifest as controversy in response to certain topics or differences in the relevance of counseling goals. At this stage, although there may be occasional confrontations in the interaction between the two, it has not escalated to strong opposition or sustained conflict.

5 = There is evident conflict and disagreement between the counselor and the client in defining and addressing the client's issues. The client may strongly oppose the counselor's viewpoints, and the counselor may shift topics, frequently interrupt, and express disagreement with the client's perspectives. At this stage, there may be clear confrontations in the interaction between both parties, leading to a compromised effectiveness of the session.

\noindent\textbf{Q4: The client and therapist have established a good understanding of the changes that would be good for the client.} 

1 = There are clear misunderstandings and disagreements between the counselor and the client in the process of change. The client may express concerns or doubts about the direction of their change, the expected outcomes of the change, or the methods of change suggested by the counselor. At this stage, more communication and guidance may be needed to build trust and understanding.

2 = The client may have doubts or uncertainties in the process of change. Although they may be taking some actions or practices, it is not clear how to achieve the expected change or the actual effectiveness of these practices. The counselor and the client need to further explore and clarify the path and expectations of change.

3 = The counselor and the client have a neutral attitude towards the process and goals of change in the conversation. Both parties may not have explicitly expressed their understanding or misunderstanding of the change. Expectations and methods of change are neither emphasized nor overlooked in the discussion, resulting in an overall lack of clear consensus or disagreement on the goals and process of counseling.

4 = Both the counselor and the client in the conversation are aware of changes that would benefit the client. This understanding may be reflected in the client's compromise on counseling goals, expressions, or discussions about the client's current situation and future expectations. Both parties are working to clarify the path and direction of change.

5 = In the counseling process, there is strong consistency and clarity between the counselor and the client regarding the client's goals and how to achieve them. They not only discuss these goals frequently and explicitly during the session but also summarize and confirm the progress and outcomes achieved at the end. The interaction and discussion at this stage align completely with the therapeutic plan.

\begin{center}
    \textbf{------ Approach ------}
\end{center}
\noindent\textbf{Q5: There is agreement about the steps taken to help improve the client’s situation.}

1 = The client directly expresses that the tasks and goals are inappropriate and generally disagrees with homework or tasks during the session. There is a disagreement between the client and the counselor regarding the approach to be taken. The client refuses to engage in tasks.

2 = The client hesitates to explore and does not follow the counselor's guidance in the change process. The client withdraws from the counselor, seeming to just "go through the motions," not engaging or focusing on the counselor or tasks. Even after some clarification by the counselor, the client still seems uncertain about the relevance of the tasks to their goals. The client appears conflicted or indifferent towards tasks in therapy and passively resists them (e.g., limited participation).

3 = There is no clear consensus or disagreement between the counselor and the client regarding therapy tasks. Both may have vague views on the significance and purpose of tasks, resulting in a neutral attitude towards participation and involvement in tasks during the session.

4 = The client shows a clear interest and involvement in therapy tasks. Whether occasional clarification is needed or not, the client participates and follows the exploration process. There is an unspoken understanding behind the tasks, leading the client to gradually acknowledge and engage in the tasks.

5 = The counselor and client strongly agree on different goals, and there is a clear disagreement on the order and solutions to issues in therapy. This inconsistency may lead the client to express strong dissatisfaction with the overall therapy process and goals, possibly mentioning the reasons for attending therapy, which may further trigger a negative reaction from the counselor. At this stage, finding common ground seems challenging, making the therapy process difficult.

\noindent\textbf{Q6: There is agreement about the usefulness of the current activity in therapy (i.e., the client is seeing new ways to look at his/her problem). }

1 = The client repeatedly argues against tasks. The client refuses to participate, claiming that it is pointless for their goals. Tension exists in the relationship between the counselor and the client, and issues are not explored.

2 = The client does not actively engage in the session tasks, although he/she may not openly question the usefulness of the tasks. The client fails to openly discuss the issues. The client may hesitate to participate in tasks but eventually engages in them. The counselor accurately conveys the reasons behind the tasks, enabling the client to understand the relevance of the tasks to their current concerns.

3 = There is no clear evidence in the communication between the counselor and the client about whether they have reached an agreement or disagreement on the client's issues. In the current interaction, there is neither a clear consensus nor an explicit conflict in opinions and feelings. Further communication and discussion may be needed to clarify their positions and expectations.

4 = The client actively participates in and is committed to therapy tasks, showing no skepticism about their effectiveness. Regardless of occasional resistance, the client engages and follows the exploration process. Both parties share a common understanding of the tasks' principles, allowing the client to gradually accept and participate in the tasks.

5 = In the counseling process, the counselor and the client have a strong and clear agreement on the client's goals and how to achieve them. They not only frequently and explicitly discuss these goals during the session but also summarize and confirm the progress and achievements at the end. The interaction and discussion at this stage align completely with the therapeutic plan.

\noindent\textbf{Q7: There is agreement on what is important for the client to work on.}

1 = There is a clear disagreement and opposition between the counselor and the client regarding the current focus. This difference may manifest as the counselor not allowing the client to shift to different topics or the client showing strong opposition during the therapy process. Their views on the direction and outcomes of therapy are entirely different.

2 = The counselor and the client have some disagreement about the content and direction of therapy, differing in the themes and time allocation to focus on during therapy.

3 = There are no clear signs of agreement or disagreement in the interaction between the counselor and the client regarding the themes or issues of therapy. Although they may engage in some exploration and communication, it is challenging to determine whether they share views on therapy themes or issues. Their reactions seem neither particularly synchronized nor explicitly conflicting.

4 = The client and the counselor respond to each other's focus and needs to some extent. They explore and accept each other's views and intentions to some degree. Although there may be some differences, they both strive to seek a common understanding and progress the therapy process.

5 = The counselor and the client are highly actively engaged in the therapy process, thoroughly exploring each other's issues and responding explicitly and continuously to each other's views and intentions. They approach therapy themes and issues with an open mindset, working together, reflecting flexibility, and demonstrating a cooperative spirit.

\noindent\textbf{Q8: The client believes that the way they are working with his/her problem is correct.}

1 = The client holds evident doubts and aversions towards the counseling process, frequently engaging in arguments with the counselor. Progress between the counselor and the client is very limited, and the time spent arguing may exceed the time dedicated to therapy. This inconsistency and questioning impact the overall therapy process.

2 = The counselor and the client sometimes have conflicting opinions, but they seem to cooperate in certain parts of the therapy process. The client expresses doubts about the therapy process or occasionally expresses concerns about certain techniques, finding other things to do during most of the counseling time.

3 = The client maintains a neutral stance toward the therapy process and methods. He/she neither explicitly expresses satisfaction nor dissatisfaction with therapy, nor does he/she clearly indicate agreement or disagreement with the therapeutic methods. During the therapy process, the client may comply at certain moments and show reservations at other times, without providing a clear evaluation of the therapy's effectiveness. This neutral attitude may stem from the client's ongoing assessment of therapy effectiveness or uncertainty about how to evaluate therapy progress.

4 = The client partially agrees with certain aspects of therapy tasks, although this agreement may not always be explicitly expressed. His/her level of involvement in the therapy process falls between simple compliance and actively providing suggestions. The client shows a certain level of agreement with the collaboration with the counselor, possibly being more actively involved in certain aspects of therapy.

5 = The client is satisfied and excited about the counselor's methods and approach to problem-solving. His/her performance in therapy is highly positive, possibly suggesting suggestions to further advance therapy tasks. Overall, the client is content with therapy work, and their interaction demonstrates a high level of cooperation and enthusiasm.

\begin{center}
    \textbf{------ Affective Bond ------}
\end{center}
\noindent\textbf{Q9: There is a mutual liking between the client and therapist.}

1 = There is evident animosity, hostility, or indifference between the counselor and the client. This may manifest in arguments, derogatory comments, or open hostility. The counselor fails to demonstrate concern for the client and may either forget important details of their life or completely disregard the client.

2 = Although there is no direct hostility between both parties, there is noticeable tension and distance in the relationship. The counselor appears indifferent or mechanical in response to the client, lacking enthusiasm. While there may not be explicit negative language, there is a lack of positive feedback and reinforcement in their interactions.

3 = There are no clear signs of warmth or coldness in the relationship between the counselor and the client. Communication lacks strong emotional feedback, and both parties seem to maintain a neutral stance. Despite engaging in communication, there is no clear expression or implication of liking or disliking each other. The relationship appears balanced without significant signs of warmth or indifference.

4 = In the majority of the sessions, the counselor and the client have positive interactions. The counselor shows enthusiasm and care for the client, frequently communicating with empathy and encouragement, exploring and understanding important details of the client's life.

5 = Throughout the therapy process, the counselor and the client consistently demonstrate a deep care for each other and provide positive feedback. The counselor not only encourages and reinforces the client's healthy behaviors but also deeply understands and cares about various aspects of the client's life, including their interests and hobbies. This profound care may lead to the client explicitly expressing gratitude and trust in the counselor. The client may also show appreciation for the counselor's care.

\noindent\textbf{Q10: The client feels confident in the therapist’s ability to help the client.}

1 = The client expresses minimal or no hope for the therapy outcomes. The client significantly questions the therapist's capabilities and may directly challenge the therapist's qualifications or understanding of the client's experiences. The client resists the therapist's suggestions, attempts at assistance, or expresses discouragement and pessimism.

2 = The client harbors doubts about the therapist, the therapy process, or the anticipated outcomes. The client may question whether the therapist truly understands their issues or doubt the interventions/homework provided during the problem-solving stages. These doubts do not come with strong opposition or hostility but noticeably impact the progress of the therapy process.

3 = The client holds a neutral stance regarding the therapist's capabilities. Throughout the therapy process, there is no clear evidence suggesting that the client has high confidence in the therapist, nor is there evidence indicating skepticism about the therapist's abilities. The client's responses and comments neither explicitly appreciate nor question the therapist's skills and capabilities.

4 = The client expresses a certain level of confidence in the therapist's abilities. This confidence may be reflected in the client's in-depth discussions on therapy topics, positive responses to the therapist's guidance, or an optimistic attitude towards resolving current counseling issues. Additionally, the client has substantial trust in the therapist's competency, possibly expressing appreciation for the effectiveness of the therapy or the therapist's abilities.

5 = The client consistently agrees with the therapist's reflections and interventions/guidance, expressing high satisfaction and appreciation for certain aspects of the therapy process or the therapist themselves. There may be multiple discussions during the therapy process highlighting the strengths of the therapy and/or the therapist.

\noindent\textbf{Q11: The client feels that the therapist appreciates him/her.}

1 = The client feels that the therapist is indifferent, inattentive, and unconcerned about his/her issues. This is expressed through explicit accusations, disdain, or other negative reactions, indicating a sense of being disregarded or misunderstood by the therapist.

2 = The client harbors some doubts about whether the therapist genuinely cares. These doubts might be indirectly expressed, such as subtle mentions or manifestations of emotions like withdrawal, displeasure, or frustration.

3 = Throughout the therapy process, there is no clear evidence of strong positive or negative reactions from the client regarding the therapist's care and support. The client neither explicitly appreciates nor expresses dissatisfaction or disregard for the therapist's sensitivity and empathetic abilities. The emotional tone of the relationship is neutral, with no apparent strong connection or distance.

4 = The therapist demonstrates a level of acceptance, warmth, and empathy towards the client, and the client perceives and responds to this caring attitude. During the therapy process, the client acknowledges to some extent the therapist's warmth and understanding.

5 = The client strongly senses the therapist's care and support, expressing gratitude for the relationship. They may praise the therapist's sensitivity and empathetic abilities, feeling comfortable and at ease for most of the therapy process.

\noindent\textbf{Q12: There is mutual trust between the client and therapist.}

1 = The client has significant mistrust towards the therapist, demonstrated by avoiding discussions on critical issues or directly expressing distrust. This mistrust hinders open communication, and the therapist may also show concerns and discomfort about the therapeutic process.

2 = There is a moderate level of mistrust between both parties, though not as intense as in the first category. The client may hesitate to share private content, and the therapist may feel a sense of uncertainty or slight discomfort regarding the therapeutic situation.

3 = There are no clear signs of trust between the therapist and client, but there are also no apparent behaviors indicating mistrust. There is a balance between trust and mistrust in their interactions, with no explicit demonstration of reliance on each other, nor clear signs of doubt or guardedness.

4 = The client is willing to disclose some personal concerns, and the therapist accepts the client's surface statements. The therapist does not overturn or interrupt the client's thoughts and maintains focus.

5 = The trust between both parties is deep enough that the client not only willingly shares deeper layers of privacy and issues but also accepts and responds to the therapist's feedback and suggestions. This level of trust enhances the overall smoothness and efficiency of the therapeutic process.

\section{Human Annotation}
\label{appendix:data_annotation}

\subsection{Human Annotators}
Since the developers are most familiar with the annotation framework and guidelines, we have selected three of them -- the postdoctoral fellow in psychology and the two NLP specialists -- to serve as annotators. Each conversation is annotated by these three individuals.

\subsection{Human Agreement}
\label{appendix: human_agreement}

Given that we plan to generalize our reliability results to any annotators with similar characteristics as the selected raters in this work, focus on the absolute agreement instead of consistency between annotators, and use the mean value of three annotators as an assessment basis, we adopt the ICC(2, k) form with two-way random effects, absolute agreement, and multiple raters. We use Pingouin package~\citep{Vallat2018pingouin} to calculate the ICC metric.

Table~\ref{tab:human_agreement} shows human agreement in evaluating working alliance across all dimensions and questions during the annotation phase.

\begin{table}[]
\centering
\scalebox{0.85}{
\begin{tabular}{cc}
\hline
\textbf{}                        & \textbf{ICC}    \\ \hline
\textbf{Q1}                      & 0.6785          \\
\textbf{Q2}                      & 0.8297          \\
\textbf{Q3}                      & 0.7337          \\
\textbf{Q4}                      & 0.7906          \\ \hline
\textit{\textbf{Goal}}           & \textit{0.7581} \\ \hline
\textbf{Q5}                      & 0.6034          \\
\textbf{Q6}                      & 0.6645          \\
\textbf{Q7}                      & 0.6055          \\
\textbf{Q8}                      & 0.7612          \\ \hline
\textit{\textbf{Approach}}       & \textit{0.6587} \\ \hline
\textbf{Q9}                      & 0.6455          \\
\textbf{Q10}                     & 0.7124          \\
\textbf{Q11}                     & 0.617           \\
\textbf{Q12}                     & 0.6241          \\ \hline
\textit{\textbf{Affective Bond}} & \textit{0.6498} \\ \hline
\textit{\textbf{Overall}}        & \textit{0.6888} \\ \hline
\end{tabular}}
\caption{Human agreement on evaluating the working alliance across all dimensions and questions.}
\label{tab:human_agreement}
\end{table}

\subsection{Data Characteristics}
\label{appendix: data_avg_scores}
Based on the annotated data, we analyze the score distribution. Table~\ref{tab:avg_scores} presents the average scores per dimension and questions along with their standard deviations in parentheses.

\begin{table}[]
\scalebox{0.82}{
\begin{tabular}{cccc}
\hline
\textbf{Dimension}                       & \textbf{Avg. Score}                  & \textbf{Question} & \textbf{Avg. Score} \\
\hline
\multirow{4}{*}{\textbf{Goal}}           & \multirow{4}{*}{3.57(0.56)}          & \textbf{Q1}       & 3.56(0.63)          \\
                                         &                                      & \textbf{Q2}       & 3.69(0.60)          \\
                                         &                                      & \textbf{Q3}       & 3.56(0.67)          \\
                                         &                                      & \textbf{Q4}       & 3.47(0.64)          \\
\hline
\multirow{4}{*}{\textbf{Approach}}       & \multirow{4}{*}{3.52(0.56)}          & \textbf{Q5}       & 3.46(0.61)          \\
                                         &                                      & \textbf{Q6}       & 3.32(0.64)          \\
                                         &                                      & \textbf{Q7}       & 3.75(0.63)          \\
                                         &                                      & \textbf{Q8}       & 3.57(0.55)          \\
\hline
\multirow{4}{*}{\textbf{Affective Bond}} & \multirow{4}{*}{\textbf{3.60(0.48)}} & \textbf{Q9}       & 3.67(0.55)          \\
                                         &                                      & \textbf{Q10}      & 3.37(0.63)          \\
                                         &                                      & \textbf{Q11}      & 3.39(0.42)          \\
                                         &                                      & \textbf{Q12}      & \textbf{3.97(0.52)} \\
\hline
\end{tabular}}
\caption{The average scores annotated on each question and dimension, with standard deviations presented in parentheses. The highest average score in each column is shown in bold.}
\label{tab:avg_scores}
\end{table}

\section{LLM Evaluation}
\label{appendix:llm_evaluation_results}

\subsection{Prompt}
\begin{figure*}
    \centering
    \scalebox{0.5}{
    \includegraphics{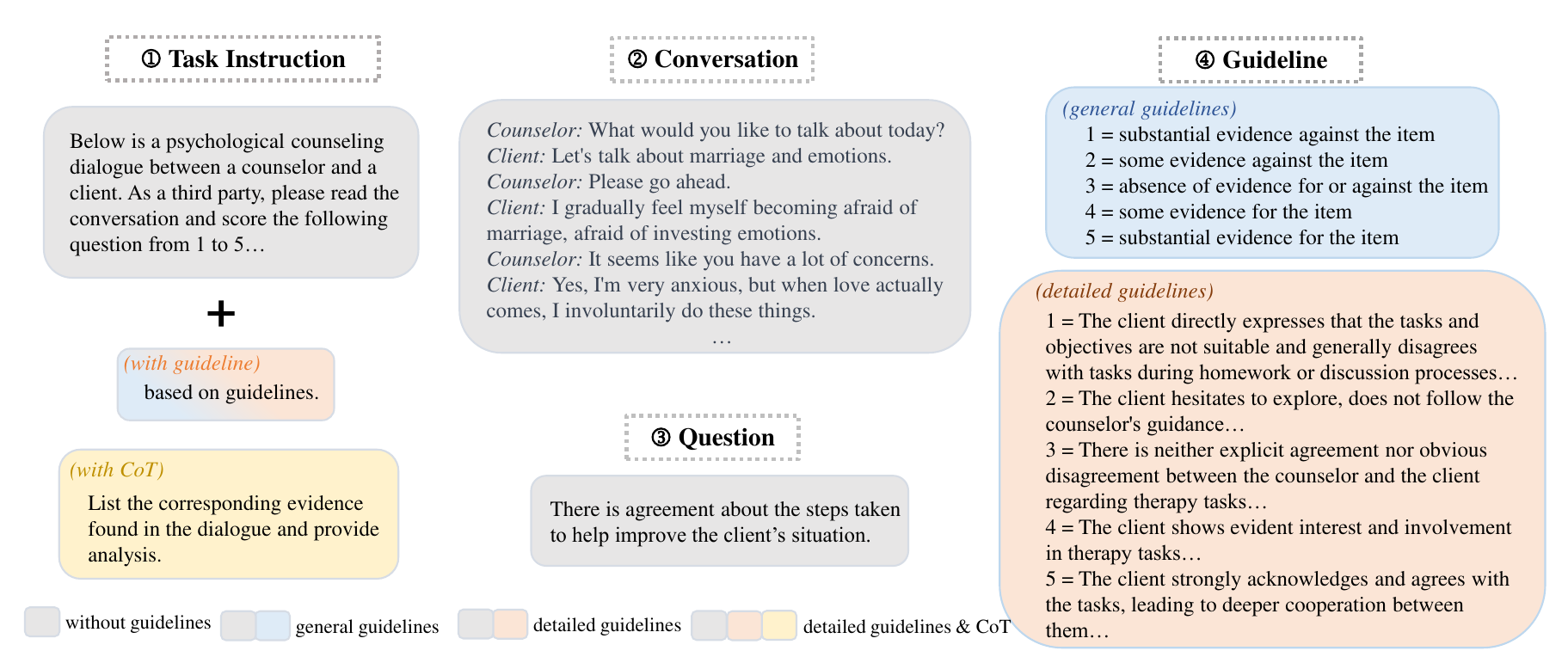}}
    \caption{Example prompts for evaluating a giving conversation across different experimental setups (i.e. with different prompt types and with/without CoT) addressing question \textit{There is agreement about the steps taken to help improve the client's situation}. General guidelines remain consistent across different questions, whereas detailed guidelines are intricately tailored to each specific question.}
    \label{fig:prompt}
\end{figure*}

Figure~\ref{fig:prompt} shows example prompts for evaluating a giving conversation across different experimental setups.

\subsection{Model Self-Agreement}
As the final annotation is determined by the average of the model's three independent annotations, we adopt the intraclass correlation coefficient with the 2-way mixed-effects model, absolute agreement definition, and the mean of $k$ measurements type as the measure of the model's self-reliability~\citep{koo2016icc_guideline, shrout1979intraclass}. Table~\ref{tab:model_agreement} presents models' intra-rater agreement on evaluating all the questions. 

\begin{table*}[]
\centering
\scalebox{0.45}{
\begin{tabular}{c|cccc|cccc|cccc|cccc}
\hline
                                 & \multicolumn{4}{c}{\textbf{ChatGPT}}                                             & \multicolumn{4}{c}{\textbf{GLM-4}}                                               & \multicolumn{4}{c}{\textbf{Claude-3}}                                            & \multicolumn{4}{c}{\textbf{GPT-4}}                                               \\
\textbf{Question}                & \textbf{No}     & \textbf{General} & \textbf{Detailed} & \textbf{Detailed + CoT} & \textbf{No}     & \textbf{General} & \textbf{Detailed} & \textbf{Detailed + CoT} & \textbf{No}     & \textbf{General} & \textbf{Detailed} & \textbf{Detailed + CoT} & \textbf{No}     & \textbf{General} & \textbf{Detailed} & \textbf{Detailed + CoT} \\
\hline
\textbf{Q1}                      & -0.2924         & 0.1989           & 0.0410            & 0.2921                  & 0.9775          & 1.0000           & 1.0000            & 0.9966                  & 0.4880          & 0.4886           & 0.8054            & 0.7779                  & 0.5359          & 0.4136           & 0.7210            & 0.7111                  \\
\textbf{Q2}                      & 0.3314          & 0.2521           & 0.5165            & 0.5972                  & 1.0000          & 1.0000           & 1.0000            & 1.0000                  & 0.7864          & 0.8379           & 0.7400            & 0.8359                  & 0.4327          & 0.6193           & 0.6884            & 0.6978                  \\
\textbf{Q3}                      & 0.0203          & -0.0130          & -0.0021           & 0.0195                  & 1.0000          & 0.9864           & 0.9957            & 0.9955                  & 0.4811          & 0.7588           & 0.5062            & 0.7061                  & 0.5935          & 0.5174           & 0.5368            & 0.6432                  \\
\textbf{Q4}                      & 0.5338          & 0.3630           & 0.5448            & 0.6179                  & 1.0000          & 1.0000           & 0.9733            & 1.0000                  & 0.8819          & 0.9278           & 0.8651            & 0.9038                  & 0.7516          & 0.8716           & 0.8500            & 0.8086                  \\
\hline
\textit{\textbf{Goal}}           & \textit{0.1483} & \textit{0.2002}  & \textit{0.2750}   & \textit{0.3816}         & \textit{0.9944} & \textit{0.9966}  & \textit{0.9922}   & \textit{0.9980}         & \textit{0.6593} & \textit{0.7533}  & \textit{0.7292}   & \textit{0.8060}         & \textit{0.5784} & \textit{0.6055}  & \textit{0.6991}   & \textit{0.7152}         \\
\hline
\textbf{Q5}                      & 0.5124          & 0.4511           & 0.5440            & 0.7828                  & 1.0000          & 0.9972           & 1.0000            & 0.9781                  & 0.8689          & 0.9058           & 0.8886            & 0.8992                  & 0.8674          & 0.8806           & 0.7648            & 0.7424                  \\
\textbf{Q6}                      & 0.3928          & 0.0686           & 0.4193            & 0.4448                  & 0.9877          & 1.0000           & 0.9879            & 1.0000                  & 0.8907          & 0.9083           & 0.7775            & 0.8921                  & 0.6768          & 0.8188           & 0.7137            & 0.6580                  \\
\textbf{Q7}                      & 0.3968          & 0.5911           & 0.3975            & 0.5755                  & 0.9933          & 0.9933           & 1.0000            & 0.9784                  & 0.7488          & 0.8373           & 0.7105            & 0.8432                  & 0.4903          & 0.7278           & 0.4286            & 0.7158                  \\
\textbf{Q8}                      & 0.6374          & 0.6196           & 0.5640            & 0.6710                  & 1.0000          & 0.9928           & 1.0000            & 0.9965                  & 0.8432          & 0.9318           & 0.8637            & 0.8518                  & 0.8279          & 0.8218           & 0.8115            & 0.7885                  \\
\hline
\textit{\textbf{Approach}}       & \textit{0.4849} & \textit{0.4326}  & \textit{0.4812}   & \textit{0.6185}         & \textit{0.9953} & \textit{0.9958}  & \textit{0.9970}   & \textit{0.9883}         & \textit{0.8379} & \textit{0.8958}  & \textit{0.8101}   & \textit{0.8716}         & \textit{0.7156} & \textit{0.8122}  & \textit{0.6796}   & \textit{0.7262}         \\
\hline
\textbf{Q9}                      & 0.7761          & 0.7614           & 0.5296            & 0.7148                  & 0.9872          & 0.9807           & 1.0000            & 1.0000                  & 0.4503          & 0.7097           & 0.8022            & 0.7404                  & 0.8439          & 0.9232           & 0.5222            & 0.5449                  \\
\textbf{Q10}                     & 0.3655          & 0.3124           & 0.5846            & 0.6225                  & 1.0000          & 0.9932           & 1.0000            & 1.0000                  & 0.8305          & 0.8414           & 0.8054            & 0.8868                  & 0.6476          & 0.7942           & 0.7920            & 0.7786                  \\
\textbf{Q11}                     & 0.7260          & 0.5660           & 0.2330            & 0.4708                  & 1.0000          & 0.9914           & 0.9948            & 0.9916                  & 0.9240          & 0.8870           & 0.8191            & 0.8027                  & 0.6716          & 0.8913           & 0.7175            & 0.8117                  \\
\textbf{Q12}                     & 0.3302          & 0.1837           & 0.4539            & 0.4418                  & 1.0000          & 0.9707           & 1.0000            & 0.9883                  & 0.6962          & 0.8538           & 0.8038            & 0.8461                  & 0.6849          & 0.6992           & 0.6781            & 0.7456                  \\
\hline
\textit{\textbf{Affective Bond}} & \textit{0.5494} & \textit{0.4559}  & \textit{0.4503}   & \textit{0.5625}         & \textit{0.9968} & \textit{0.9840}  & \textit{0.9987}   & \textit{0.9950}         & \textit{0.7252} & \textit{0.8230}  & \textit{0.8076}   & \textit{0.8190}         & \textit{0.7120} & \textit{0.8270}  & \textit{0.6774}   & \textit{0.7202}         \\
\hline
\textit{\textbf{Overall}}        & \textit{0.3942} & \textit{0.3629}  & \textit{0.4022}   & \textit{0.5209}         & \textit{0.9955} & \textit{0.9921}  & \textit{0.9960}   & \textit{0.9938}         & \textit{0.7408} & \textit{0.8240}  & \textit{0.7823}   & \textit{0.8322}         & \textit{0.6687} & \textit{0.7482}  & \textit{0.6854}   & \textit{0.7205}   \\
\hline
\end{tabular}}
\caption{The intrarater reliability of models in evaluating each question and dimension across different experimental settings.}
\label{tab:model_agreement}
\end{table*}

\subsection{Alignment with Human Evaluations}
The alignment between LLMs and human evaluations are presented in  Table~\ref{tab:pearson_correlation_5_likert}.

\begin{table*}[]
\centering
\scalebox{0.4}{
\begin{tabular}{c|llll|llll|llll|llll}
\hline
\multicolumn{1}{l|}{}            & \multicolumn{4}{c|}{\textbf{ChatGPT}}                                            & \multicolumn{4}{c|}{\textbf{GLM-4}}                                                & \multicolumn{4}{c|}{\textbf{Claude-3}}                                                     & \multicolumn{4}{c}{\textbf{GPT-4}}                                                       \\
\textbf{Question}                & \textbf{No}     & \textbf{General} & \textbf{Detailed} & \textbf{Detailed + CoT} & \textbf{No}     & \textbf{General}   & \textbf{Detailed} & \textbf{Detailed + CoT} & \textbf{No}     & \textbf{General}   & \textbf{Detailed}        & \textbf{Detailed + CoT}  & \textbf{No}     & \textbf{General} & \textbf{Detailed}        & \textbf{Detailed + CoT}  \\ \hline
\textbf{Q1}                      & -0.0462         & 0.1743           & 0.1014            & 0.1139                  & 0.2818*         & 0.4359***          & 0.4186***         & 0.4469***               & 0.3752***       & 0.1473             & 0.3657***                & 0.5503***                & 0.2406*         & 0.3012**         & \textbf{0.5379***}       & 0.4292***                \\
\textbf{Q2}                      & 0.2415*         & 0.0303           & 0.2978**          & 0.2877*                 & 0.3840***       & 0.4491***          & 0.4236***         & 0.4447***               & 0.4293***       & 0.2663*            & \textbf{0.4976***}       & 0.3994***                & 0.3423**        & 0.3698***        & 0.4712***                & 0.5379***                \\
\textbf{Q3}                      & -0.1578         & -0.0171          & 0.1453            & 0.2430*                 & 0.2614*         & 0.146              & 0.4721***         & 0.4650***               & 0.1758          & 0.3229**           & \textbf{0.4987***}       & 0.4249***                & 0.3869***       & 0.2920**         & 0.4907***                & 0.4510***                \\
\textbf{Q4}                      & 0.2904**        & 0.1192           & 0.4497***         & 0.157                   & 0.3477**        & 0.4582***          & 0.3593**          & 0.2841*                 & 0.5482***       & 0.5551***          & 0.5180***                & 0.4460***                & 0.4667***       & 0.3651***        & 0.4919***                & \textbf{0.5569***}       \\ \hline
\textit{\textbf{Goal}}           & \textit{0.082}  & \textit{0.0767}  & \textit{0.2486}   & \textit{0.2004}         & \textit{0.3187} & \textit{0.3723}    & \textit{0.4184}   & \textit{0.4102}         & \textit{0.3821} & \textit{0.3229}    & \textit{0.47}            & \textit{0.4552}          & \textit{0.3591} & \textit{0.332}   & \textit{\textbf{0.4979}} & \textit{0.4937}          \\ \hline
\textbf{Q5}                      & 0.4624***       & 0.2061           & 0.4070***         & 0.4222***               & 0.4253***       & 0.5058***          & 0.4738***         & 0.4610***               & 0.5542***       & \textbf{0.6485***} & 0.5048***                & 0.6088***                & 0.5710***       & 0.6423***        & 0.5618***                & 0.6025***                \\
\textbf{Q6}                      & 0.4033***       & 0.2998**         & 0.3290**          & 0.3599**                & 0.5716***       & \textbf{0.6798***} & 0.4378***         & 0.6558***               & 0.6160***       & 0.5891***          & 0.4183***                & 0.5949***                & 0.5237***       & 0.6190***        & 0.5065***                & 0.5371***                \\
\textbf{Q7}                      & 0.13            & 0.214            & 0.3924***         & 0.3392**                & 0.3982***       & 0.4350***          & 0.4141***         & 0.4145***               & 0.4069***       & 0.3815***          & 0.3612**                 & 0.4283***                & 0.3764***       & 0.2921**         & \textbf{0.5341***}       & 0.4924***                \\
\textbf{Q8}                      & 0.4179***       & 0.3464**         & 0.2058            & 0.3233**                & 0.2516*         & 0.3172**           & 0.3949***         & 0.4703***               & 0.3081**        & 0.2703*            & 0.5180***                & \textbf{0.6114***}       & 0.2439*         & 0.2532*          & 0.5898***                & 0.5472***                \\ \hline
\textit{\textbf{Approach}}       & \textit{0.3534} & \textit{0.2666}  & \textit{0.3336}   & \textit{0.3612}         & \textit{0.4117} & \textit{0.4844}    & \textit{0.4301}   & \textit{0.5004}         & \textit{0.4713} & \textit{0.4724}    & \textit{0.4506}          & \textit{\textbf{0.5608}} & \textit{0.4288} & \textit{0.4516}  & \textit{0.548}           & \textit{0.5448}          \\ \hline
\textbf{Q9}                      & 0.185           & 0.3062**         & 0.3577**          & 0.3752***               & 0.2229*         & 0.1725             & 0.4801***         & \textbf{0.5555***}      & -0.1563         & -0.0277            & 0.2851*                  & 0.3027**                 & 0.0106          & 0.1325           & 0.2337*                  & 0.3086**                 \\
\textbf{Q10}                     & 0.4433***       & 0.3144**         & 0.3352**          & 0.4273***               & 0.5401***       & 0.5507***          & 0.4512***         & 0.4520***               & 0.6269***       & \textbf{0.6839***} & 0.5957***                & 0.5420***                & 0.5164***       & 0.6339***        & 0.5114***                & 0.4520***                \\
\textbf{Q11}                     & 0.4943***       & 0.3920***        & 0.4633***         & 0.4570***               & 0.5256***       & 0.5250***          & 0.5705***         & 0.5834***               & 0.4463***       & 0.5250***          & 0.5528***                & 0.4975***                & 0.4994***       & 0.3874***        & \textbf{0.6113***}       & 0.6103***                \\
\textbf{Q12}                     & 0.2651*         & 0.1914           & 0.2507*           & 0.3892***               & 0.4981***       & 0.4717***          & 0.4552***         & 0.4079***               & 0.4853***       & 0.4035***          & \textbf{0.5762***}       & 0.5727***                & 0.4506***       & 0.4305***        & 0.4101***                & 0.4960***                \\ \hline
\textit{\textbf{Affective Bond}} & \textit{0.3469} & \textit{0.301}   & \textit{0.3517}   & \textit{0.4122}         & \textit{0.4466} & \textit{0.43}      & \textit{0.4893}   & \textit{0.4997}         & \textit{0.3506} & \textit{0.3962}    & \textit{\textbf{0.5024}} & \textit{0.4787}          & \textit{0.3693} & \textit{0.3961}  & \textit{0.4417}          & \textit{0.4667}          \\ \hline
\textit{\textbf{Overall}}        & \textit{0.2608} & \textit{0.2148}  & \textit{0.3113}   & \textit{0.3246}         & \textit{0.3924} & \textit{0.4289}    & \textit{0.4459}   & \textit{0.4701}         & \textit{0.4013} & \textit{0.3971}    & \textit{0.4743}          & \textit{0.4982}          & \textit{0.3857} & \textit{0.3933}  & \textit{0.4959}          & \textit{\textbf{0.5018}} \\ \hline
\end{tabular}}
\caption{Pearson correlation between human and model annotations on each dimension and question. Statistic significance levels for individual question correlations are denoted by $***p \textless 0.001$, $**p \textless 0.01$, and $*p \textless 0.05$. The overall and dimension-specific correlations are calculated as the averages of the correlations on corresponding questions.}
\label{tab:pearson_correlation_5_likert}
\end{table*}

\section{LLM-based Insights}

\subsection{Counselors' Abilities in Establishing Relationships with Clients}
\begin{figure}
    \centering
    \scalebox{0.36}{
    \includegraphics{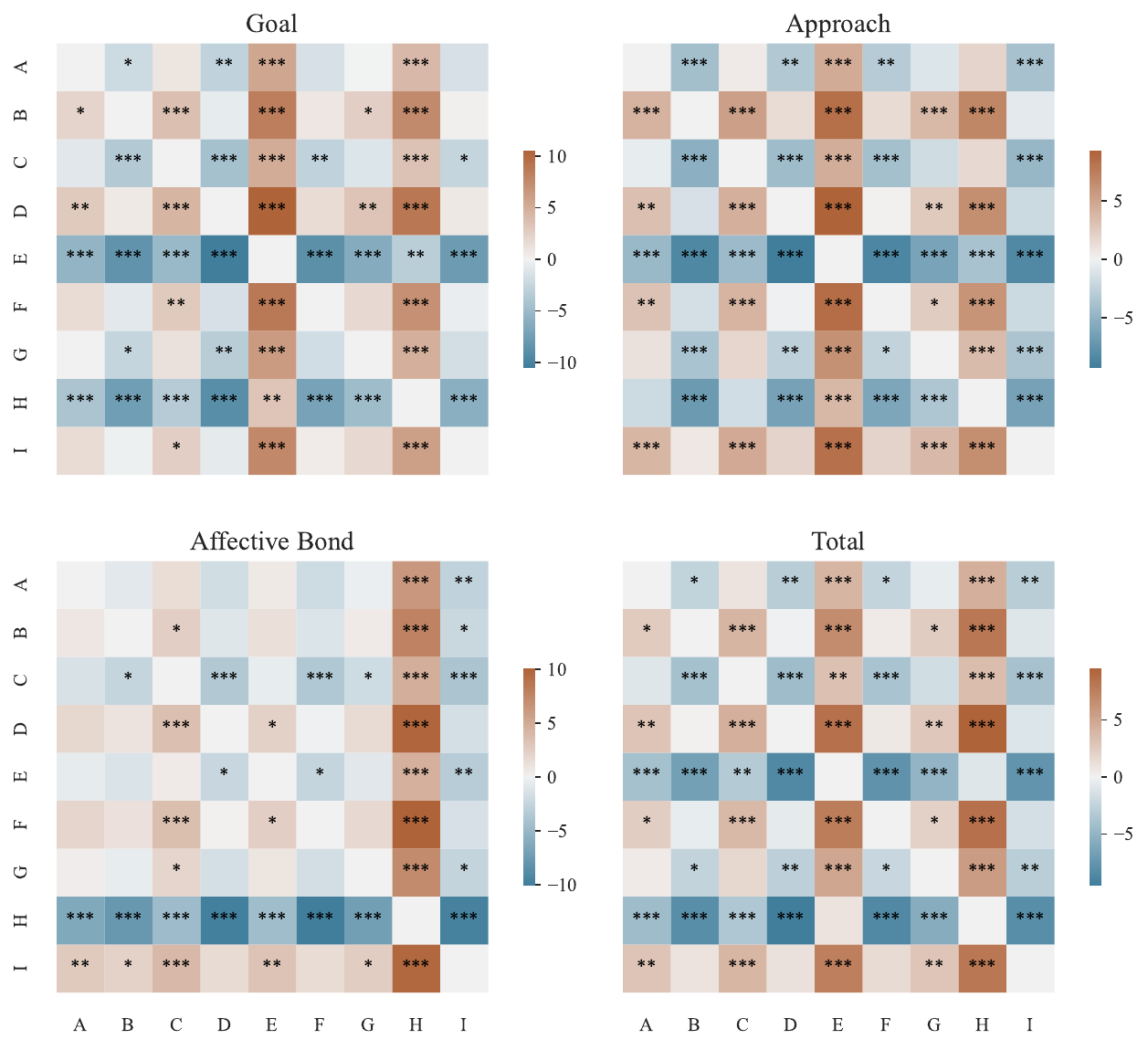}}
    \caption{The heatmap results of pairwise t-tests on the working alliance scores of counselors across all counseling sessions with their clients, where each element in the heatmap represents the t value with significance. *$p$ < 0.05, **$p$ < 0.01, ***$p$ < 0.001.}
    \label{fig:ttest_counselor_score}
\end{figure}

Figure~\ref{fig:ttest_counselor_score} shows the heatmap results of pairwise t-tests on the working alliance scores of counselors across all counseling sessions with their clients.

\label{appendix:counselor_ability}

\subsection{Correlation between Working Alliance and Outcomes}
\label{appendix: wa_outcomes}
The Outcome Rating Scale (ORS) is designed to assess change in clients following psychological interventions, demonstrating adequate validity, solid reliability, and high feasibility~\citep{miller2003outcome}. In our study, clients are required to fill out the Outcome Rating Scale (ORS) before each counseling session to report their conditions following the previous counseling session, aiming to indicate the effectiveness of the previous counseling.

The English version of the introduction and scale items are: \textit{Kindly utilize a scale ranging from 0 to 100 to assess your overall quality of life for the past week, encompassing today as well. Here, 0 signifies the lowest point, while 100 symbolizes the highest, with higher scores denoting more favorable conditions. Please input your score into the provided box: 
1. Individual Physical and Mental Well-being; 2. Interpersonal Relationships (Family or any Intimate Relationships); 3. Social Life (Work, School, Friends); 4. Overall Condition.}

Table~\ref{tab:wa_ors} demonstrates the Pearson correlation coefficients between the dimensions of working alliance and the dimensions of the ORS.

\begin{table}[]
\centering
\scalebox{0.65}{
\begin{tabular}{ccccc}
\hline
                        & \textbf{\makecell[c]{Phy. \& Men.}} & \textbf{\makecell[c]{Relationship}} & \textbf{Social Life} & \textbf{Overall} \\ \hline
\textbf{Goal}           & 0.20***                     & 0.29***                             & 0.27***              & 0.27***          \\
\textbf{Approach}       & 0.21***                     & 0.30***                             & 0.26***              & 0.27***          \\
\textbf{Affective Bond} & 0.29***                     & 0.35***                             & 0.32***              & 0.34***          \\
\textbf{Total}          & 0.25***                     & 0.34***                             & 0.31***              & 0.32***          \\ \hline
\end{tabular}}
\caption{The Pearson correlation coefficients between each dimension of working alliance and ORS.}
\label{tab:wa_ors}
\end{table}

\subsection{Implication for LLM-based Feedback}
\label{appendix:llm-based_feedback}

We request counselors E and H to evaluate the LLM-based feedback. They are tasked with rating the following three questions for each feedback: (1) I believe this can help me better understand the alliance between myself and the client; (2) I think this can inspire me to focus on certain aspects to enhance the alliance between myself and the client; and (3) I am willing to adjust my counseling strategies in future sessions based on this feedback. Their rating results are presented in Table~\ref{tab:rating_llm_feedback}. The results indicate that they harbor a positive perspective regarding the effectiveness of LLM-based feedback in aiding them to better comprehend their relationships with clients and offering potential directions for improvement. Additionally, they express a willingness to adjust their strategies based on the feedback.

\begin{table}[]
\centering
\scalebox{0.64}{
\begin{tabular}{ccccc}
\hline
\textbf{No.} & \textbf{Question}  & \textbf{E} & \textbf{H} & \textbf{Avg.} \\
\hline
1 & \makecell[l]{I believe this can help me better understand \\ the alliance between myself and the client.}                      & 3.56                 & 3.30                 & 3.43          \\
\hline
2 & \makecell[l]{I think this can inspire me to focus on \\ certain aspects to enhance the alliance \\ between myself and the client.} & 3.78                 & 3.20                 & 3.49          \\
\hline
3 & \makecell[l]{I am willing to adjust my counseling strategies \\ in future sessions based on this feedback.}                     & 3.78                 & 3.70                 & 3.74   \\
\hline
\end{tabular}}
\caption{The assessment results of counselors E and H on the LLM-based feedback.}
\label{tab:rating_llm_feedback}
\end{table}

\section{The Consent Form and User Services Agreement}
\label{consent form}

Below are the English translation of consent forms and user services agreement used in the current work, the original documents are in Mandarin Chinese. Every client gave their consent to attend the online text-based psycho-counseling on our counseling platform and agreed to data usage for the current work. 

\subsection{Consent Form}

\noindent{Dear clients,}

Thank you for your trust. Before we formally begin the counselings, there are some relevant matters that need to be communicated to you, so that the consultation can proceed smoothly and effectively. This agreement is the basic framework to ensure the normal conduct of the psychological consultation process. Please read it carefully and tick the box at the bottom to indicate your agreement. If you have any questions, please raise them with your counselor after the counselings.

1. Duration and Frequency of Consultation: Psychological consultations require regular sessions, each typically lasting 50 minutes. The frequency and total duration of the consultations will be jointly determined by you and your counselor based on the nature of your psychological distress and personal needs.

2. Confidentiality and Exceptions to Confidentiality: In general, your counselor will keep the information you provide confidential, including case records, test materials, letters, recordings, videos, and other materials, all of which are considered professional information and are stored under strict confidentiality to prevent public disclosure in any public setting. However, there are exceptions to confidentiality in the following cases, and relevant individuals and institutions will be notified: 

1) Violation of relevant laws (e.g., if you pose a danger to others; suspicion of child or elder abuse or abuse of someone dependent on you for care, etc.) 

2) If your situation endangers your own safety (e.g., suicide, self-harm, mental illness, severe depression, etc.), we will notify your relatives or guardians when necessary and consult your opinion to ensure your safety. 

3) Counselors need to receive supervision during their work. Counselors will discuss parts of the consultation content and visitor information in personal supervision and case discussions. Privacy information unrelated to the consultation, such as personal names and regions, will be anonymized; supervisors and case discussion members are also bound by the aforementioned confidentiality rules. If there is a need to publicly release or publish consultation details, the visitor's written consent must be obtained first.

3. Adjusting Consultation Times: If you wish to adjust your consultation time, please do so at least 24 hours in advance on the platform. Adjustments cannot be made if the time limit is exceeded.

4. Handling of Lateness: You may enter the counseling from the start of the scheduled appointment until it ends, but the end time of the consultation will not be extended due to your lateness. If you are late and do not log in to start the consultation by the service end time, the consultation will be considered expired, and the consultation fee will not be refunded.

5. Responsibilities of the Clients: 
During the consultation process, visitors need to:

1) Attend and participate in the consultation sessions;

2) Express and share their thoughts and feelings as much as possible during the consultation;

3) Seriously reflect on their own expressions, the counselor's responses, and the interaction process between the two.

6. Responsibilities of the Counselor: 
Counselors need to:

1) Arrange a suitable consultation schedule for both parties;

2) Strive to guide visitors towards an understanding of themselves and their current situation, and help them better deal with the various difficulties and life events they are facing;

3) Regularly participate in professional learning and case discussions to ensure their competence in counseling work with visitors;

4) Be aware of their limitations as a counselor and discuss ending the consultation or referrals with the visitor in a timely manner if the consultation is ineffective or unsuccessful.

7. Duration and Frequency of Consultation:

1) Psychological consultations are regular sessions, typically 50 minutes each, once a week. Changes to the interval and frequency will be determined based on the nature of your psychological issues and personal needs.

2) Consultation sessions will start and end on time. Flexibility in timing will not exceed 5 minutes.

8. Emergency Consultation: In urgent situations, you may make a temporary appointment or call the local crisis intervention hotline.

9. Crisis Intervention Measures: In the event that you are experiencing severe psychological stress with thoughts of suicide and impulses, it is necessary to discuss potential risks and coping strategies with a counselor. This includes how to access local support resources and techniques for self-regulation. Due to the limitations of remote counseling, counselors may be unable to work with visitors at high risk of suicide. In cases of intense suicidal urges or self-destructive behavior, counselors are obligated to discuss referral to appropriate assistance agencies. (National 24-Hour Suicide Intervention Hotline: 4001619995)

10. Physical symptoms and psychological symptoms often interact, and if necessary, we may discuss the need for consultation and treatment in medical institutions during counseling. Additionally, medication can be beneficial at the appropriate time in alleviating both physical and mental issues. Throughout the treatment process, based on your specific situation, the counselor may recommend relevant laboratory and instrumental examinations, providing detailed explanations as needed.

11. Psychological counseling and therapy are complex processes that may require coordination, continuous goal adjustment, or referrals and other interventions during the course.

12. Voluntary Withdrawal: You have the right to terminate your counseling at any time, but it is recommended to discuss and carefully conclude with your counselor before formal withdrawal.

13. If there are other research and teaching matters that require your participation, your counselor will inform you and negotiate with you to sign an additional written agreement.

14. During the period of the consultation work, if there is a need to adjust or modify the agreement, both parties can propose it during the consultation. After thorough discussion and agreement, corresponding changes will be made.

\noindent\textbf{Remote/Online Counseling Additional Matters:}

When conducting online counseling, identity verification is required. For this purpose, you need to provide some materials (such as personal information, current situation, etc.) to complete this process.

For situations not suitable for online counseling, such as suicidal or homicidal thoughts, life-threatening circumstances, a history of suicidal, abusive, or violent tendencies, hallucinations, and substance or alcohol abuse, it is recommended to consider face-to-face counseling or alternative intervention methods.

Considering the potential impact on the counseling relationship, please refrain from recording audio or video during the counseling process. If there is a genuine need for such recordings, it should be discussed thoroughly and agreed upon by both parties.

The smooth conduct of online counseling depends on stable network conditions, communication devices, and a disturbance-free room. Please ensure that you are adequately prepared before starting online counseling. Additionally, be psychologically prepared for unforeseen events such as network interruptions during online counseling.

[ ] I fully understand and agree to the above terms.

\subsection{Informed Consent Form in the User Services Agreement}

VI. Informed Consent

\noindent{6.1 To protect your rights, please read and agree before activating the dialogue service of this application: Users agree to accept the online text counseling or venting services (hereinafter referred to as the service) provided by this application based on my confusions. Users understand that the current service provided by this application is AI-assisted psychological counseling/venting, with real human counselors also providing services. Users need to understand that the online text venting/counseling service is an internet-based form of instant psychological confusion resolution and psychological knowledge popularization service. This service is provided in Chinese. Users need to understand that the service content includes support and help for psychological confusions (including, but not limited to: emotional issues, relationship issues, family relations, interpersonal relationships, personal growth, career development, etc.). Although it is difficult to guarantee a complete improvement in psychological conditions and resolution of confusions, we serve you with the attitude of "sometimes curing, often helping, always comforting". Users need to understand that during the service process: conversations will involve the user's physiological/psychological health and emotional state among other related information. Users have the right to privacy in the venting/counseling service, and the personal information disclosed by users will, in principle, be kept strictly confidential. At the same time, the user's right to privacy is protected and restricted by national laws in terms of content and scope. Users need to understand, based on national laws, there are exceptions to the principle of confidentiality, including but not limited to the following situations:}

1) When the service seeker or others are preparing or in the process of engaging in actions that endanger the safety of themselves or others' person or property;

2) When the service seeker may endanger others (such as in cases of contagious diseases);

3) When the information disclosed by the service seeker involves a minor being or about to be sexually abused;

4) When the service seeker or others are preparing or in the process of engaging in actions that endanger national security or public safety;

5) In cases where data is anonymized for discussions, consultations, or when receiving supervision and training among consulting members;

6) In cases where data is anonymized for scientific research.

7) When disclosure is required by law.

\noindent{6.2 Users must agree that for the aforementioned non-confidential situations, for the fundamental reason of protecting the rights of the user or related individuals, we may disclose information to the minimal extent necessary and only within the necessary scope of personnel. Furthermore, users must understand that since the counseling service is conducted over the internet, although we strive to protect users' privacy to the greatest extent, it is difficult to avoid the possibility of personal information being leaked due to internet security vulnerabilities, technical failures, or unauthorized access. Users must understand that under the following conditions, we are unable to provide effective venting/counseling services, and it is necessary to seek professional offline treatment or counseling services:}

1. Having thoughts or plans of suicide;

2. Having thoughts or plans of harming oneself or others;

3. Having any psychiatric disorder diagnosed by a hospital;

4. Meeting the diagnostic criteria for any psychiatric disorder.

Users need to understand that if the physiological, psychological, mental state, and behavior plans described or reflected in their information meet any of the above criteria, we cannot continue to provide services to them, and may suggest seeking professional offline treatment or counseling services. Users must understand that this application provides support and help for psychological confusions (including but not limited to: emotional issues, relationship issues, family relations, interpersonal relationships, personal growth, career development, etc.), but there still exist some services that are difficult to provide:

1) Crisis intervention for suicide or other harmful behaviors;

2) Diagnosis and treatment of psychiatric disorders;

3) Specific advice on the use of psychiatric medications;

4) Dealing with severe psychological trauma;

5) Providing specific resources or information for careers, academics, etc.;

6) Providing views on social phenomena and interpretations of policies;

7) Interpretation of dreams (e.g., explaining the meaning of dreams, why certain people or things appear in dreams, etc.).

8) To answer psychological confusions not related to myself (for example, those of my friends, family, online friends, etc.). 

Users need to understand that when the described situation exceeds our service scope (which does not include the aforementioned 8 types), we cannot meet their needs. Users need to understand the potential benefits and risks of internet-based text venting/counseling services. The benefits include, but are not limited to, being able to access services more conveniently without the need to travel to a designated location. And, although the risks are small, users still understand that there may be potential risks. These risks include, but are not limited to: due to possibly insufficient information provided by the user, the services received may not fully resolve the user's confusions or improve the user's psychological state; due to possible technical failures or other unforeseen reasons, the user may not receive timely analysis and advice for their psychological confusions. Users must agree that when the application provides services, it follows the laws and regulations of mainland China, not the laws and regulations of the user's location. The above informed consent remains effective during the user's single or multiple uses of the service.

\noindent{6.3. I agree to convert the collected psychological counseling dialogue text data into digital and graphical forms for use in non-profit academic cooperation, academic conferences, journal publications, and other academic activities by certified third-party academic institutions (*1).}

(*1) Certified third-party academic institutions refer to universities and research institutes officially recognized by the state, and researchers working within them have undergone formal academic training.

\end{CJK*}
\end{document}